\documentclass[final,5p,times]{elsarticle}

\usepackage{hyperref}

\journal{ArXiv}

\usepackage{graphicx}
\usepackage{graphics}
\usepackage{float}
\usepackage[caption = false]{subfig}
\usepackage{amsfonts}
\usepackage{amsmath}
\usepackage{textcomp}
\usepackage[table]{xcolor}    

\def\eg{\emph{e.g. }}

\def\etal{\emph{et al. }}
\def\ie{\emph{i.e. }}

\definecolor{tableGray}{gray}{0.90}

\DeclareMathOperator*{\argmin}{arg\,min}

\begin{document}

\begin{frontmatter}

\title{The challenge of simultaneous object detection and pose estimation: a comparative study}

\author{Daniel O\~noro-Rubio, Roberto J. L\'opez-Sastre, Carolina Redondo-Cabrera and Pedro Gil-Jim\'enez}
\address{GRAM, University of Alcal\'a, Alcal\'a de Henares, 28805, Spain}



\ead{daniel.onoro@edu.uah.es, robertoj.lopez@uah.es, carolina.redondoc@edu.uah.es and pedro.gil@uah.es}



\begin{abstract}
Detecting objects and estimating their pose remains as one of the major challenges of the computer vision research community. There exists a compromise between localizing the objects and estimating their viewpoints. The detector ideally needs to be view-invariant, while the pose estimation process should be able to generalize towards the category-level. This work is an exploration of using deep learning models for solving both problems simultaneously. For doing so, we propose three novel deep learning architectures, which are able to perform a joint detection and pose estimation, where we gradually decouple the two tasks. We also investigate whether the pose estimation problem should be solved as a classification or regression problem, being this still an open question in the computer vision community. We detail a comparative analysis of all our solutions and the methods that currently define the state of the art for this problem. We use PASCAL3D+ and ObjectNet3D datasets to present the thorough experimental evaluation and main results. With the proposed models we achieve the state-of-the-art performance in both datasets.
\end{abstract}


\end{frontmatter}

\section{Introduction}
Over the last decades, the category-level object detection problem has drawn considerable attention. As a result, much progress has been realized, leaded mainly by international challenges and benchmarking datasets, such as the PASCAL VOC Challenges \cite{everingham2010} or the ImageNet dataset \cite{deng2009imagenet}. Nevertheless, researchers soon identified the importance of not only localizing the objects, but also estimating their poses or viewpoints, \eg \cite{Thomas2006,Savarese2008,Lopez-Sastre2011,Yingze-Bao2011}. This new capability results fundamental to enable a true interaction with the world and its objects. For instance, a robot which merely knows the location of a cup but that cannot find its handle, will not be able to grasp it. In the end, the robotic solution needs to know a viewpoint estimation of the object to facilitate the inference of the visual affordance for the object. Also, in the augmented reality field, to localize and estimate the viewpoint of the objects, is a crucial feature in order to project a realistic hologram, for instance.

\begin{figure}[t]
\includegraphics[width=0.95\columnwidth]{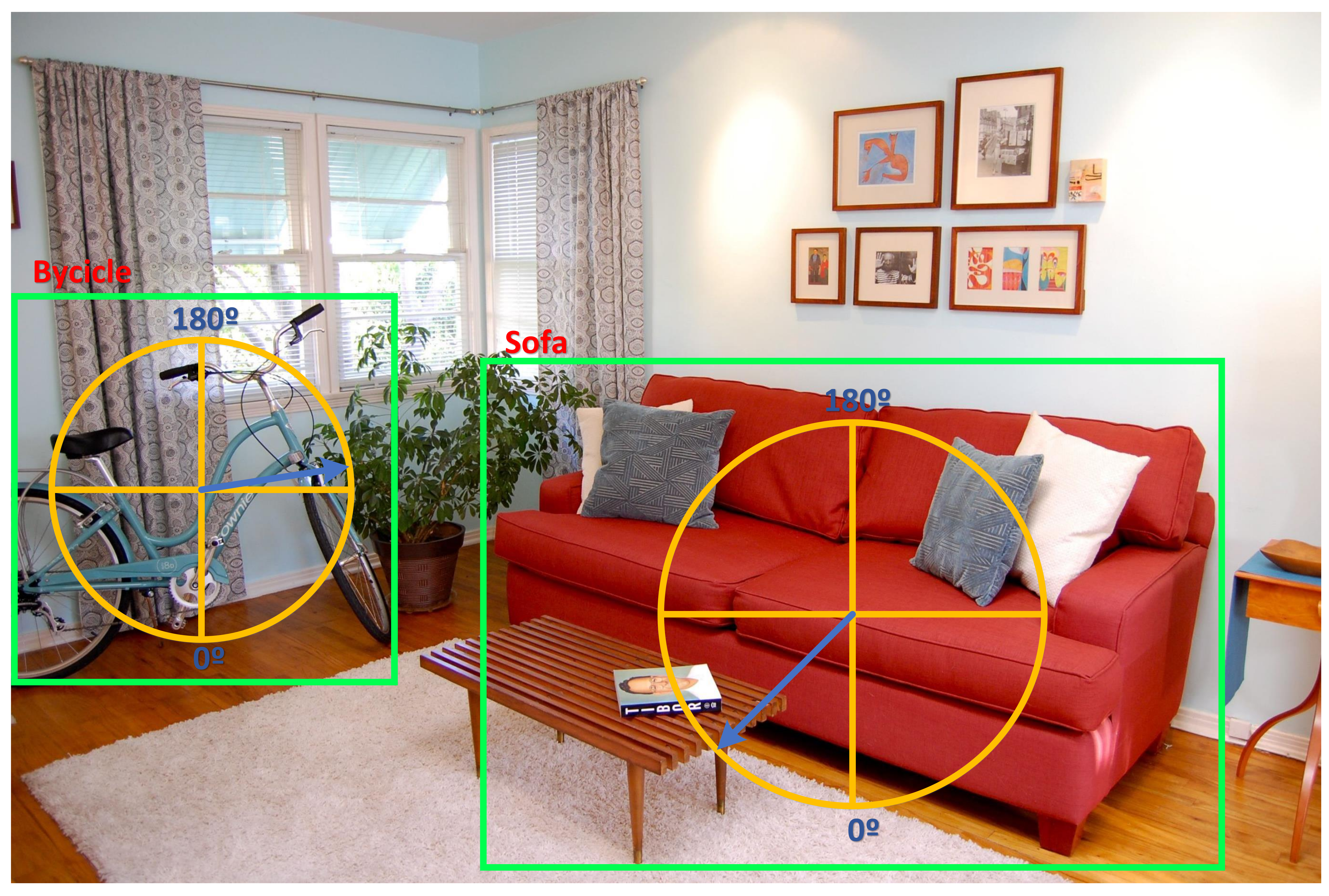} 
\label{fig:problem_intuition}
\caption{Object category detection and pose estimation example. In the image, the sofa and the bicycle are localized by the green bounding boxes. The blue arrow inside the yellow circles shows the azimuth angles of the objects, which is a form of viewpoint annotation.}
\end{figure}

Technically, given an image, these models can localize the objects, predicting their associated bounding boxes, and are also able to estimate the relative pose of the object instances in the scene with respect to the camera. Figure \ref{fig:problem_intuition} shows an example, where the viewpoint of the object is encoded using just the azimuth angle. In the image, the target objects are the sofa and the bicycle. Their locations are depicted by their bounding boxes (in green), and their azimuth angles are represented by the blue arrow inside the yellow circle. 

The computer vision community rapidly detected the necessity of providing the appropriate annotated datasets, in order to experimentally validate the object detection and pose estimations approaches. To date, several datasets have been released. Some examples are: 3D Object categories \cite{Savarese2008}, EPFL Multi-view car \cite{Ozuysal2009}, ICARO \cite{icaro}, PASCAL3D+  \cite{xiang2014} or ObjectNet3D \cite{xiang2016}.

Thanks to these datasets, multiple models have been experimentally evaluated. It is particularly interesting to observe how all the published approaches can be classified in two groups. In the first one, we find those models that decouple both problems (\eg \cite{Tulsiani_vpsKps_2015,Glasner2012,Redondo-Cabrera2015}), making first a location of the object, to later estimate its pose. In the second group we identify the approaches that solve both tasks simultaneously (\eg \cite{xiang2014,Pepik2012,MassaBMVC2016}), because they understand that to carry out a correct location requires a good estimation of the pose, and vice versa.

But the discrepancies do not end here. Unlike the problem of object detection, where the metric for the experimental evaluation is clear, being this the mean Average Precision (mAP) defined in the PASCAL VOC Challenge, for the problem of object detection and pose estimation, multiple metrics have been adopted. This is motivated by the fact that not all the models understand the viewpoint estimation problem in the same way. Some solutions, \ie the discrete approaches, consider that this is a classification problem, when others, \ie the continuous models, understand the pose as a continuous variable, whose estimation must be approached by solving a regression problem.

This article is an attempt to provide a comparative study where these issues can be addressed. The main contributions of this work are as follows: 
\begin{itemize}
 \item We introduce three novel deep learning architectures for the problem of simultaneous object detection and pose estimation. Our models seek to perform a joint detection and pose estimation, trained fully end-to-end. We start with a model that fully integrates the tasks of object localization and object pose estimation. Then, we present two architectures that gradually decouple both tasks, proposing a final deep network where the integration is minimal. All our solutions are detailed in Sections \ref{sec:learning_model} and \ref{sec:analyzed_models}.
 \item All our architectures have been carefully designed to be able to treat the pose estimation problem from a continuous or from a discrete perspective. We simply need to change the loss functions used during learning. This is detailed in Section \ref{sec:loss_pose}. Therefore, in our experiments, we carefully compare the performance of these two families of methods, reporting results using four different loss functions. Therefore, this paper aims to shed some light on which perspective is more appropriate, keeping the network architecture fixed.
 \item Thanks to the proposed models, we are able to offer an experimental evaluation (see Section \ref{sec:experiments}) designed to carefully analyze how coupled the detection and pose estimation tasks are, being this our final contribution. We also bring a detailed comparison with all the solutions that establish the state-of-the-art for the problem of object category detection and pose estimation. We carefully analyze all the models using two publicly available datasets: PASCAL3D+ \cite{xiang2014} and ObjectNet3D \cite{xiang2016}.
\end{itemize}
\section{Related Work}
\label{sec:rel_work}

Object category detection and viewpoint estimation is a growing research field. Several are the methods that have contributed to improve the state of the art. Like we just have said, we can organize in two groups all the approaches in the literature. 

In the first one, we find those models that understand that these two tasks, \ie object localization and pose estimation, must be solved separately \cite{Tulsiani_vpsKps_2015,Glasner2012,Redondo-Cabrera2015}. The second group consists of the models where the detection and the viewpoint estimation are fully coupled \cite{xiang2014,Pepik2012,MassaBMVC2016,Redondo-Cabrera2014}.

Within these two groups, one must note that while some models solve the pose estimation as a classification problem, \ie the discrete approaches \cite{MassaBMVC2016,TulsianiposeInductionTCM15}, others treat the viewpoint estimation as a regression problem, \ie the  continuous solutions \cite{Glasner2012,Redondo-Cabrera2014,Fenzi2013}. 

In this paper, we introduce three novel deep learning architectures for the problem of \emph{joint} object detection and pose estimation. They all are extensions for the excellent Faster R-CNN object detection model \cite{ren2015fasterrcnn}. We have designed them to gradually decouple the object localization and pose estimation tasks. Our models significantly differ from previous deep learning based approaches for the same tasks. For instance, if we consider the work of Tulsiani \etal \cite{TulsianiposeInductionTCM15}, we observe that their solution is based on a detector (using the R-CNN \cite{girshick2014}), followed by a pose classification network, fully decoupling both tasks. On the contrary, all our architectures are trained fully end-to-end, performing a joint detection and viewpoint estimation. Moreover, the deep architectures implemented are different. Massa \etal \cite{MassaBMVC2016} also propose a \emph{joint} model. However, their approach is completely different. They base their design on the Fast R-CNN detector \cite{girshickICCV15fastrcnn}. Technically, they modify the Fast R-CNN output to provide the detections based on an accumulative sum of scores that is provided by the pose classification for each object category. In a different manner, our solutions are based on the Faster R-CNN, which is a distinct architecture. Moreover, in our work we explore not only a modification of the output of the networks, but multiple architecture designs where we can gradually separate the branches of the network dedicated to the object localization and the viewpoint estimation tasks. 

Finally, this paper offers a detailed comparative study of solutions for the joint object \emph{detection} and pose estimation problem. The study included in \cite{elhoseiny2016} focus on the different problem of object \emph{classification} and pose estimation,\ie they do not consider the object localization task.

\section{Simultaneous detection and pose estimation models}
\label{sec:model}

\begin{figure}[t]
\centering
\subfloat[Single-path architecture.]{\includegraphics[width=0.95\linewidth]{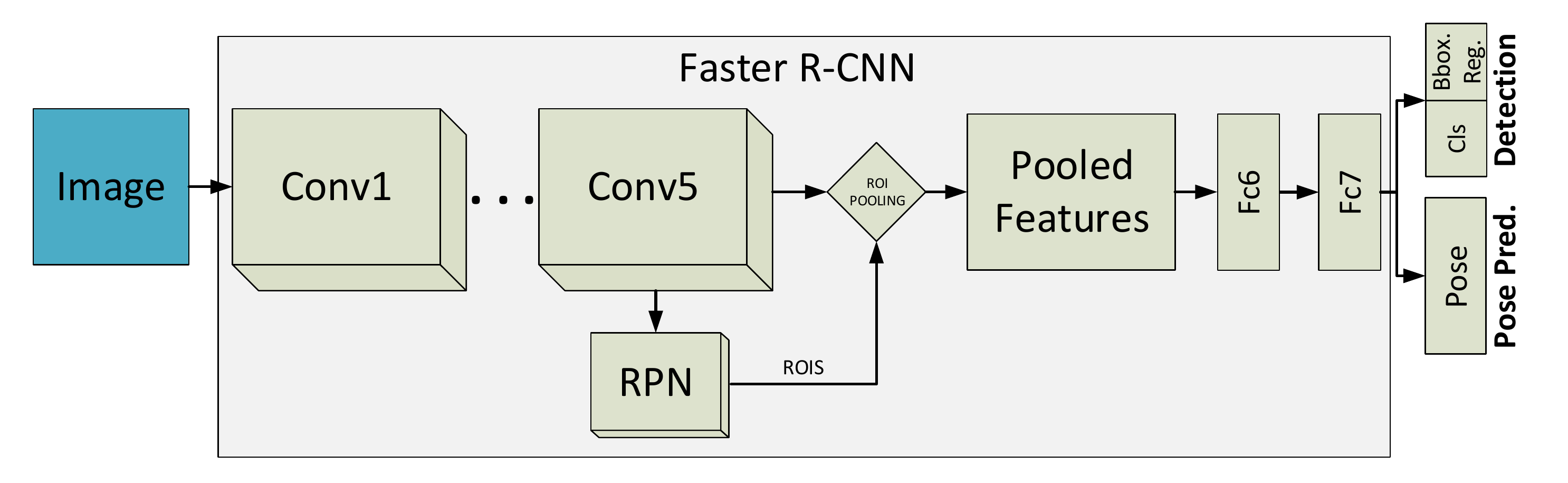} \label{fig:analized_models:baseline}}\
\subfloat[Specific-path architecture.]{\includegraphics[width=0.95\linewidth]{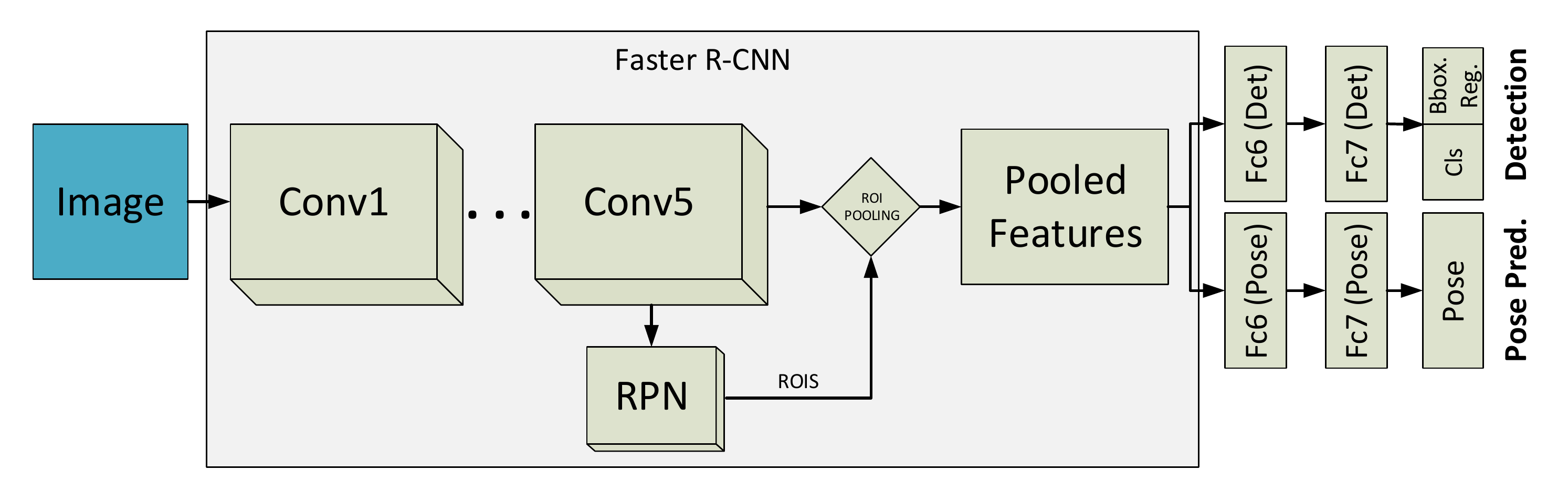} \label{fig:analized_models:pathspec}}\  
\subfloat[Specific-network architecture.]{\includegraphics[width=0.95\linewidth]{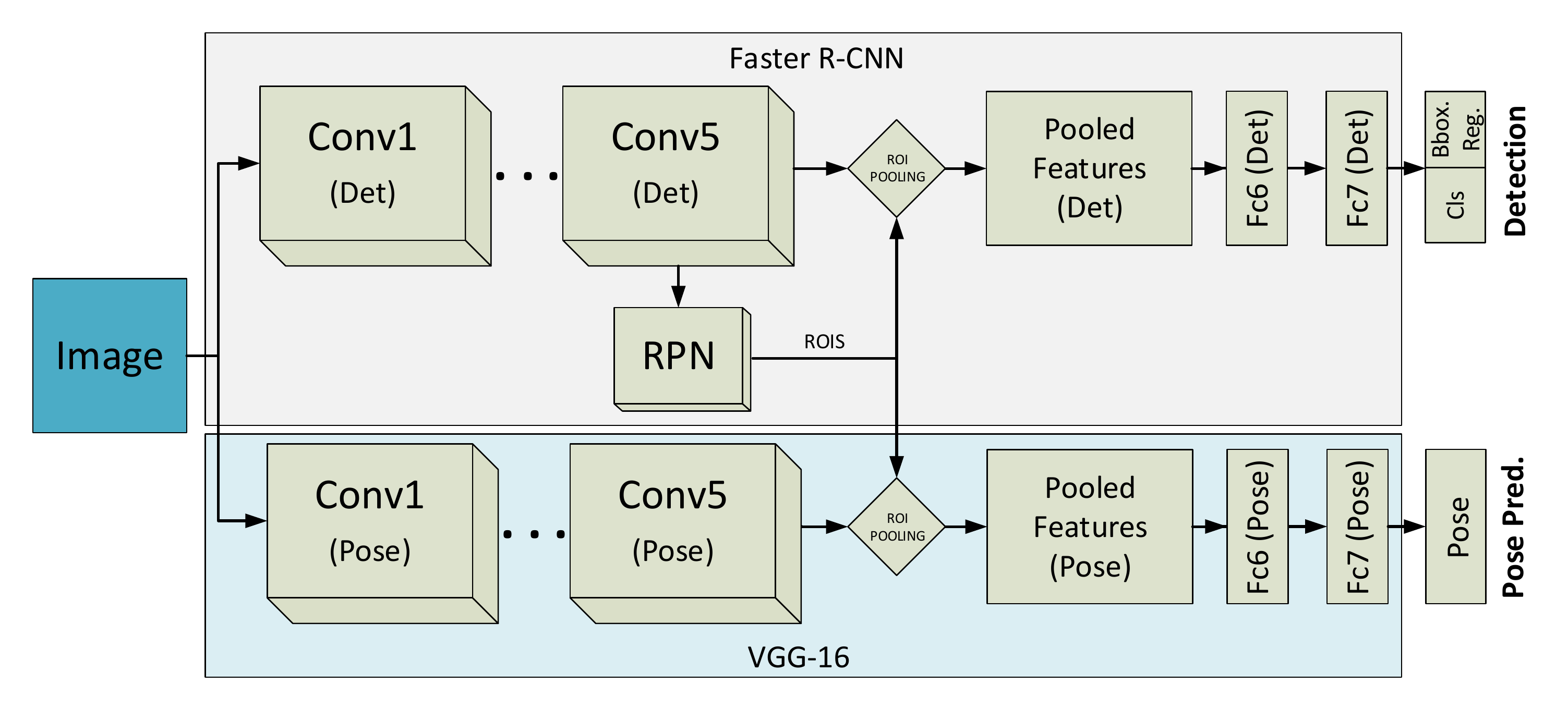} \label{fig:analized_models:netspec}} 
\caption{Proposed deep learning architectures for simultaneous object detection and pose estimation.}
\label{fig:analized_models}
\end{figure}

In the following section, we formulate the learning problem for a joint detection and pose estimation. Then, we detail the proposed architectures, named: single-path, specific-path, and specific-network (Figure \ref{fig:analized_models} shows an overview of all our designs). Technically, they all are extensions for the Faster R-CNN approach \cite{ren2015fasterrcnn}. Finally, we provide a detailed analysis of the loss functions used in our experimental evaluation.

\subsection{Learning model for simultaneous detection and pose estimation}
\label{sec:learning_model}

Our goal is to learn a strong visual representation that allows the models to: localize the objects, classify them and estimate their viewpoint with respect to the camera. Furthermore, we consider an \emph{in the wild} setting where multiple objects of a variety of categories appear in real-world scenarios, with a considerable variability on the background, and where occlusions and truncations are the rule rather than the exception.

Therefore, the supervised learning process starts from a training set $S = \{(x_{i}, t_i)\}_{i = 1}^{N}$, where $N$ is the number of training samples. For each sample $i$ in the dataset, $x_i \in X$ represents the input image, and $t_{i} \in T$, with $t_{i} = (y_i, \beta_i, \phi_i)$, encodes the annotations for the three tasks to solve: classification ($y_i$), object localization ($\beta_i$) and pose estimation ($\phi_i$). $y_i \in Y$ with $Y = [1,2,\dots, C, C+1]$ describes the object class, being $C$ the total number of object categories. Category $C+1$ is used to consider a generic background class. $\beta_i \in \mathbb{R}^4$ represents the bounding box localization of a particular object within image $x_{i}$. Finally, $\phi_i \in \mathbb{R}^3$ encodes the 3D viewpoint annotation for a particular object with respect to the camera position as a tuple of azimuth, elevation and zenith angles.

We propose to learn a convolutional neural network (CNN) \cite{LeCun1990} for simultaneous object detection and pose estimation. Technically, these CNNs are a combination of three main features which let the model achieve a sort of invariance with respect to imaging conditions: local receptive fields, shared convolutional weights, and spatial pooling. Each unit in a layer receives inputs from a set of units located in a small neighborhood of the previous layer. In the forward pass of a CNN, each output feature is computed by the convolution of the input feature from the previous layer. Therefore, these deep networks can be thought of as the composition of a number of convolutional structure functions, which transform the input image to feature maps that are used to solve the target tasks.

For the particular problem of simultaneous object detection and viewpoint estimation, our CNN prediction $\hat{t}$ should be expressed as follows,

\begin{equation}
\hat{t}_{\boldsymbol{\theta},W} = F_{W} \circ \mathbf{z_{\theta}}(x_i) \, .
\label{eq:prediction}
\end{equation}

$\mathbf{z_{\theta}}: X \rightarrow \mathbb{R}^D$ represents the $D$-dimensional feature mapping that the network performs to the input images. Technically, it consists in the transformation of the input image $x_i$ into a feature that is used to feed the output layers of our models. We encode in $\boldsymbol{\theta}$ the trainable weights of the deep architecture that allow the network to perform the mapping. In our solutions, the weights in $\boldsymbol{\theta}$ define the hidden layers that are shared by all the tasks that the deep network needs to solve.

$F_{W}$ corresponds the set of functions of the output layers. They take as input the deep feature map $\mathbf{z_{\theta}}(x_i)$. For the problem considered in this paper, our set of functions must address three different tasks: classification ($y$), object localization ($\beta$) and viewpoint estimation ($\phi$). Therefore, $F_{W} = (f^y_{W^y},f^\beta_{W^\beta},f^\phi_{W^\phi})$. $f^y_{W^y}$ with weights $W^y$ produces the prediction for the object category, \ie $\hat{t}^y$. $f^\beta_{W^\beta}$ predicts the object location $\hat{t}^\beta$. Finally, $f^\phi_{W^\phi}$ is in charge of the prediction of the viewpoint $\hat{t}^\phi$.

According to the prediction model detailed in Equation \ref{eq:prediction}, we define the following objective function to learn our multi-task neural network:

\begin{equation}
\argmin_{\boldsymbol{\theta},W}  \mathcal{L}(\boldsymbol{\theta},W,S) \, ,
\label{eq:objective_function}
\end{equation}
where the loss function follows the equation,

\begin{equation}
\mathcal{L}(\boldsymbol{\theta},W,S) = \lambda_1 \mathcal{L}_y(\boldsymbol{\theta},W^y,S) + \lambda_{2} \mathcal{L_{\beta}}(\boldsymbol{\theta},W^\beta,S) + \lambda_{3} \mathcal{L}_{\phi}(\boldsymbol{\theta},W^\phi,S) \, .
\label{eq:global_loss_function}
\end{equation}

$\lambda_{i}$ for $i \in (1,2,3)$ corresponds to the scalar value that controls the importance of a particular loss during training. For the classification loss $\mathcal{L}_y$ we use a categorical cross-entropy function. A simple Euclidean loss is used for the object localization task loss $\mathcal{L_{\beta}}$. Finally, for the pose estimation loss $\mathcal{L}_{\phi}$ multiple options are considered. We detail them in Section \ref{sec:loss_pose}.

\subsection{The proposed architectures}
\label{sec:analyzed_models}

\subsubsection{Single-path architecture}
\label{sec:single_model}

Our first deep network design is the \emph{single-path} architecture. It offers a natural extension of the Faster R-CNN model for the problem of simultaneous object detection and pose estimation. Technically, we simply add an extra output layer in order to predict the viewpoint of the object.

To understand the extension proposed, we proceed with a description of the original Faster R-CNN pipeline. As it is shown in Figure \ref{fig:analized_models:baseline}, the Faster R-CNN consists of three stages. The first stage is performed by the convolutional layers. An input image passes through the convolutional network, to be transformed into a deep feature map. The second stage is represented by the Region Proposal Network (RPN), which serves as an ``attention'' mechanism during learning. Technically, it is a fully convolutional (sub)network, which takes an image feature map as input, and outputs a set of rectangular object proposals, with their corresponding objectness scores. To go into details, this RPN takes the feature map obtained from the last convolutional layer (\eg convolution 5 in a VGG16-based architecture), and  adds a new convolutional layer which is in charge of learning to generate regions of interest (ROIs). In the third stage, these ROIs are used for pooling those features that are passed to the last two fully-connected (FC) layers. Finally, the responses coming from the last FC layer are used by the model: 1) to classify the ROIs into background or object; and 2) to perform a final bounding box regression for a fine-grained localization of the object. In Figure \ref{fig:analized_models:baseline} we represent these two tasks with the blocks named as ``Cls'' (for classification) and ``Bbox. Reg.'' (for the bounding box regression). Technically, the ``Cls'' module is implemented with a softmax layer, and the ``Bbox. Reg.'' layer is a linear regressor for the four coordinates that define a bounding box.

In order to evaluate the capability of the Faster R-CNN for the task of pose estimation, guaranteeing a minimal intervention in the model architecture, we propose the \textit{single-path} extension. It consists in the incorporation of an additional output layer (see box ``Pose'' in Figure \ref{fig:analized_models:baseline}), connected to the last FC layer as well. The objective of this layer is to cast a prediction for the viewpoint, and to measure the loss for this task, propagating the appropriate gradients to the rest of the network during learning. 

For training this \textit{single-path} model, we solve the objective loss function of Equation \ref{eq:objective_function}. We give the same weight to each task, \ie $\lambda_{1} = \lambda_{2} = \lambda_{3} = 1$.
Note that at this point, we do not specify whether the viewpoint estimation will be considered as a classification or regression problem. In this sense, different loss functions will be considered an evaluated in the experiments, in order to attain a high level of understanding of the simultaneous detection and pose estimation problem.

\subsubsection{Specific-path architecture}
\label{sec:specific_path}

The \emph{specific-path} is our second approach. Our objective with this architecture is to explore the consequences of a slightly separation of the pose estimation task from the object class detection, learning \emph{specific deep features} for each task. 

As it is shown in Figure \ref{fig:analized_models:pathspec}, the extension we propose for this second approach consists in adding two independent FC layers, which are directly connected to the pose estimation layer. Note that we do not change the rest of the architecture, \ie both the initial convolutional layers and the RPN module are shared. The pooled features are used to feed the two groups of FC layers that form two types of features: one for the object detection task, and the other for the viewpoint estimation. Therefore, during training, each network FC path learns its specific features based on its gradients, while the rest of layers learn a shared representation.

The model is learned solving the objective function shown in Equation \ref{eq:objective_function}. For the detection path, $\lambda_{1} = \lambda_{2} = 1$, and $\lambda_{3} =0$. For the pose path we solve the Equation \ref{eq:objective_function} getting rid of the object classification and bounding box regression losses, \ie $\lambda_{1}=\lambda_{2}=0$ and $\lambda_{3} = 1$.

\subsubsection{Specific-network architecture}
\label{sec:specific_network}

With our third architecture, named \emph{specific-network}, we attempt to separate as much as possible the detection and pose estimation tasks within the same architecture. The key idea of this design is to provide a model with two networks that can be fully specialized in their respective tasks, while they are learned simultaneously and end-to-end.

Consequently, as it is shown in Figure \ref{fig:analized_models:netspec}, we design a model made of two independent networks: the \emph{detection network} and the \emph{pose network}. The \emph{detection network} is in charge of fully performing the object localization task, as in the original design of the Faster R-CNN.

The \emph{pose network} must focus on the viewpoint estimation task, without any influence of the detection objective. Therefore, this network has now its own initial convolutional layers. To align the detection and pose estimation, the \emph{pose network} receives the ROIs generated by the RPN module of the \emph{detection network}. Technically, an input image is forwarded simultaneously into both convolutional networks. The second stage of the Faster R-CNN, \ie the generation of ROIs by the RPN, occurs in the detection network only. These ROIs are shared with the \emph{pose network}. Finally, each network pools its own features from the generated ROIs, feeds its FC layers with these features, and produces its corresponding outputs.

Overall, we have an architecture with two specialized networks, that are synchronized to solve the object detection and pose estimation tasks in a single pass. 

For learning this model we follow the same procedure as for the \emph{specific-path}.  We train our \emph{detection network} to solve the Equation \ref{eq:objective_function} where $\lambda_{3} = 0$ and  $\lambda_{1}=\lambda_{2}=1$. The \emph{pose network} is solved just for the pose problem, hence, $\lambda_{1}=\lambda_{2}=0$ and $\lambda_{3}=1$. The main difference with respect the \emph{specific-path} model is that there are no shared features, so each network is fully specialized to solve its corresponding task.

\subsubsection{Why have we chosen these designs?}
\label{sec:discussion_designs}

All our architectures are extensions of the Faster R-CNN approach \cite{ren2015fasterrcnn}. Originally, the Faster R-CNN architecture was proposed to address the problem of object detection only. This model has systematically prevailing on all the detection benchmarks (\eg PASCAL VOC \cite{everingham2010}, COCO \cite{Lin2014} and ILSVRC detection \cite{deng2009imagenet}), where leading results are obtained by Faster R-CNN based models, albeit with deeper features (\eg using deep residual networks \cite{He2016}). So, following a simple performance criterion, we believe that the Faster R-CNN with its excellent results is a good choice.

Our second criterion for the selection of this Faster R-CNN architecture is related with the main objective of our research: propose and evaluate solutions for the problem of \emph{simultaneous} object detection and viewpoint estimation. Note that we neither address the problem of pose estimation in a classification setup in isolation (\eg \cite{elhoseiny2016}, where the object localization problem is not considered), nor decouple the object detection and pose estimation tasks (\eg \cite{TulsianiposeInductionTCM15}). Our models seek to perform a \emph{joint} detection and pose estimation, trained fully end-to-end, and the Faster R-CNN architecture is an ideal candidate to extend. All our solutions perform a direct pooling of regions of interests in the images from the internal RPN of the Faster R-CNN. This way, we do not need to use any external process to hypothesize bounding boxes (\eg Selective Search \cite{uijlings2013}), hence performing a truly end-to-end simultaneous object detection and pose estimation model, where the weights of the fully convolutional RPN learn to predict object bounds and objectness scores at each position, to maximize not only the object detection accuracy, but also the viewpoint estimation performance.

Finally, we want to discuss our main arguments for the concrete extensions proposed in our architectures. Traditionally, the computer vision community working on the problem of pose estimation for object categories has been divided into two groups. Those that understand that the tasks of localizing objects and estimating their poses are decoupled tasks (\eg \cite{TulsianiposeInductionTCM15,Glasner2012,Fenzi2013,Redondo-Cabrera2015}), and those that advocate for jointly solving both tasks (\eg \cite{MassaBMVC2016,Su_2015_ICCV,Redondo-Cabrera2014,Pepik2012}). The architectures proposed in this paper move from a \emph{fully} integration of both tasks, \ie in the \emph{single-path}, towards the \emph{specific-network} model, where the integration is minimal. In this way, we can design an experimental evaluation to thoroughly analyze how coupled the detection and pose estimation tasks are. Moreover, all our experiments are carried on publicly available dataset which have been designed for the problem of detection and viewpoint estimation, therefore a direct comparison with previous methods that define the state of the art is also possible.

\subsection{Loss functions for pose estimation}
\label{sec:loss_pose}

Unlike the well-defined object detection task, the viewpoint estimation problem has been traditionally considered from two different perspectives: the continuous and the discrete. Most methods in the literature adopt the discrete formulation. That is, they understand the pose estimation as a classification problem, relying on a coarse quantization of the poses for their multi-view object detectors (\eg \cite{Tulsiani_vpsKps_2015,Pepik2012,Su_2015_ICCV}). Only a few approaches consider that the pose estimation of categories is ultimately a continuous problem, \ie a regression problem (\eg \cite{Redondo-Cabrera2014,Fenzi2013,Beyer2015BiternionNets}). In this paper, all our architectures are evaluated considering these two perspectives for the viewpoint estimation. 

When we want our models to consider discrete outputs for the pose estimation (the ``Pose'' layer in Figure  \ref{fig:analized_models}), we integrate the following categorical cross-entropy loss function in Equation \ref{eq:objective_function}:

\begin{equation}
\mathcal{L}_{\phi}(\boldsymbol{\theta},W^{\phi},S) = - \frac{1}{N}\sum_{i=1}^{N} \log \left(\sigma_{l^{\phi}_i}( f^\phi_{W^\phi} \circ \mathbf{z_{\theta}}(x_i)\right) \, ,
\label{eq:softmax}
\end{equation}
where $N$ is the number of samples, and $\sigma_{l^{\phi}}$ is the softmax function for the label $l^{\phi}_i$.

When the pose estimation is considered from the continuous perspective, multiple adequate regression loss functions can be integrated. For all them, it is fundamental to deal with the circularity of the viewpoint. Therefore, we first represent the orientation angles as points on a unit circle by the following transformation, $\mathbf{p}(\alpha) = (\sin(\alpha), \cos(\alpha))$, $\mathbf{p}(\alpha) \in \mathbb{R}^2$.

Probably, the simplest way to train the pose regressor is by using an Euclidean loss, as follows:

\begin{equation}
\mathcal{L}_{\phi}(\boldsymbol{\theta},W^{\phi},S) = \frac{1}{2N}\sum_{i=1}^{N} \left( \mathbf{p}\left( l^{\phi}_i\right) - \mathbf{p}\left(f^\phi_{W^\phi} \circ \mathbf{z_{\theta}}(x_i)\right)\right)^2 \, .
\label{eq:euclideanloss}
\end{equation}

A popular alternative to the Euclidean loss, is the Huber loss function, 
\begin{equation}
\resizebox{0.95\columnwidth}{!}{
$
\mathcal{L}_{\phi}(\boldsymbol{\theta},W^{\phi},S) = \frac{1}{N}\sum_{i=1}^{N} \left\{
\begin{matrix}
\frac{1}{2} \left(\mathbf{p}\left(l^{\phi}_i\right) - \mathbf{p}\left(f^\phi_{W^\phi} \circ \mathbf{z_{\theta}}(x_i)\right)\right)^2 & \text{if } \left | \mathbf{p}\left(l^{\phi}_i\right) - \mathbf{p}\left(f^\phi_{W^\phi} \circ \mathbf{z_{\theta}}(x_i)\right) \right | \leq \delta , \\ 
\delta  \left | \mathbf{p}\left(l^{\phi}_i\right) - \mathbf{p}\left(f^\phi_{W^\phi} \circ \mathbf{z_{\theta}}(x_i)\right) \right | - \frac{1}{2} \delta^2 &  \text{otherwise} 
\end{matrix}  \right. \, .
$
} 
\label{eq:huberloss}
\end{equation}

The advantage of this loss is that it tends to be more robust to outliers than the Euclidean loss.

Finally, we propose to also use the continuous cyclic cosine cost function, which is widely used in the natural language processing literature \cite{singhal2001}. It is defined as follows,

\begin{equation}
\mathcal{L}_{\phi}(\boldsymbol{\theta},W^{\phi},S) = \frac{1}{N}\sum_{i=1}^{N} \left( 1 - \frac{ \mathbf{p}(l^{\phi}_i) \mathbf{p}(f^\phi_{W^\phi} \circ \mathbf{z_{\theta}}(x_i))} { \left \| \mathbf{p}(l^{\phi}_i) \right \| \left \| \mathbf{p}(f^\phi_{W^\phi} \circ \mathbf{z_{\theta}}(x_i)) \right \| } \right) .
\label{eq:cyccosloss}
\end{equation}

\section{Experiments}
\label{sec:experiments}

\subsection{Implementation details}
\label{sec:implementation_details}
To perform our experiments, we have implemented all our models and loss functions using the deep learning framework Caffe \cite{jia2014caffe}. The optimization is done by using the Stochastic Gradient Descent algorithm, with: a momentum of 0.9; a weight decay of 0.0005; and a learning rate of 0.001. The learning rate of the output layer for the pose estimation has been multiplied by a factor of 0.01, so as to guarantee that the network properly converges. We publicly release all our implementations\footnote{The link to download all the models and software to reproduce the results will be inserted once the paper gets accepted.}.

We follow the standard procedure of the Faster R-CNN \cite{ren2015fasterrcnn} for training the models in an end-to-end fashion. This way, for each training iteration, just one image is taken and passed through the first set of convolutions. In a second step, a collection of 128 region proposals is generated. These regions are used to build the batch to feed the last set of FC layers. This batch contains 32 samples of foreground samples and 96 samples of background.

For the experimental evaluation, we use two publicly available datasets, which have been especially designed for the evaluation of object detection and pose estimation models: PASCAL3D+ \cite{xiang2014} and ObjectNet3D \cite{xiang2016}. We strictly follow the experimental setup described in these datasets. In the following sections, more details are provided, as well as a thorough analysis of the results and main conclusions obtained.

\subsection{Results in the PASCAL3D+ dataset}
\label{sec:experiment_pascal3d+}

\begin{figure*}[t]
\centering
\includegraphics[width=0.75\linewidth]{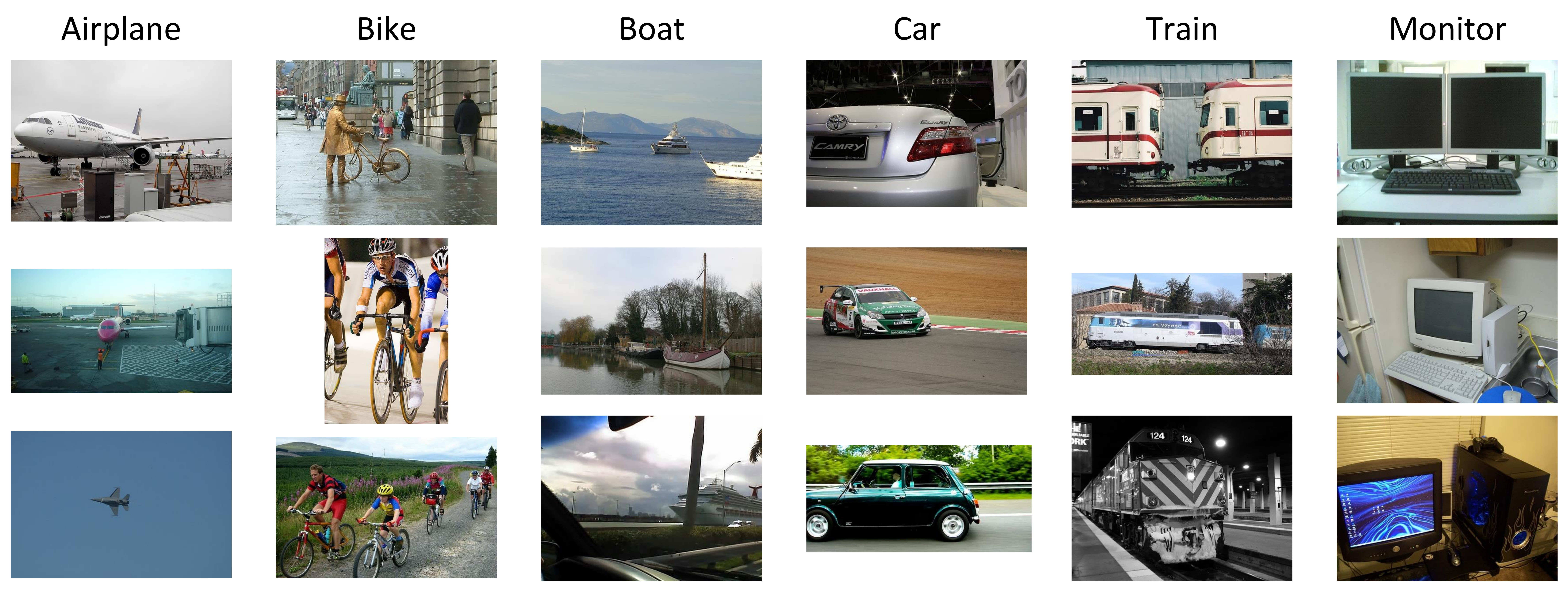} 
\caption{Some images of the PASCAL3D+ dataset.}
\label{fig:pascal3d_sample}
\end{figure*}

PASCAL3D+ \cite{xiang2014} dataset is one of the largest and most challenging datasets for the problem of object detection and pose estimation. Technically, it consists of: 1) the images and annotations of the 12 rigid object categories provided with the PASCAL VOC 2012 dataset \cite{everingham2010}; and 2) an additional set of 22,394 images taken from the ImageNet \cite{deng2009imagenet} dataset, for the same 12 categories. On average, it has more than 3000 instances per object category. The test set has 5823 images directly inherited from the PASCAL VOC 2012 test subset. Figure \ref{fig:pascal3d_sample} shows some examples of images. One can clearly observe that the images provided contain objects ``in the wild''. The standard PASCAL VOC annotation for all the objects (\ie category label and bounding box), has been extended to provide a precise 3D pose. This has been done performing a manual alignment of the objects in the images with 3D CAD models. This way, azimuth, elevation and distance from the camera pose in 3D are provided for each object.

For our analysis, we follow the official experimental setup of the PASCAL3D+ \cite{xiang2014}. The evaluation metric for the object detection and pose estimation is the Average Viewpoint Precision (AVP). This AVP is similar to the Average Precision (AP) for object detection. To compute the AVP, every output of the detector is considered to be correct if and only if the bounding box overlap with the ground truth annotation is larger than 50\% \emph{and} the viewpoint estimation for the azimuth angle is correct. When we consider a discrete space for the viewpoint, the viewpoint estimation is correct if it coincides with the ground truth azimuth label. On the contrary, if the viewpoint belongs to a continuous space, then, two viewpoint labels are correct if the distance between them is smaller than a fixed threshold of $\frac{2 \pi}{v}$, where $v$ is the number of views.

\subsubsection{Network initialization analysis}

One of the most common practices in deep learning consists in initializing a deep network architecture with the weights of a model pre-trained in a big dataset, such as ImageNet \cite{deng2009imagenet}, and then start a fine tunning process for a specific task, typically using a different dataset.

For our problem of joint object detection and pose estimation, we also follow this popular recipe. In a nutshell, we fine tune our networks in the PASCAL3D+ dataset, using for the initialization of the weights two pre-trained models: the original VGG16 model \cite{ simonyan2014} trained for the ImageNet dataset, and the Faster R-CNN model \cite{ren2015fasterrcnn} using only the training set of the PASCAL VOC 2012 dataset. Note that the validation set of the original PASCAL VOC 2012 is now the test set proposed in the PASCAL3D+, therefore, we do not allow the Faster R-CNN to be pre-trained on it.For the rest of model weights that are not covered by the pre-trained models, we basically follow a standard random initialization.

Here we simply want to explore what initialization procedure is the best option. Therefore, for this preliminary experiment, we just use our first architecture, the \textit{Single-path}. The pose estimation is considered as a classification problem, using 360 discrete bins, and we employ the cross-entropy loss defined in Eq. \ref{eq:softmax}.

\begin{table}[h]
\centering
\resizebox{\columnwidth}{!}{%
\begin{tabular}{lccccc}
\hline 
Init. strategy & mAP & mAVP 4 & mAVP 8 & mAVP 16 & mAVP 24 \\ \hline
\hline
\multicolumn{1}{|l|}{ImageNet} & \multicolumn{1}{c|}{49.5} & \multicolumn{1}{c|}{37.6} & \multicolumn{1}{c|}{32.0} & \multicolumn{1}{c|}{\textbf{24.6}} & \multicolumn{1}{c|}{\textbf{20.2}}\\ \hline
\multicolumn{1}{|l|}{PASCAL VOC 2012} & \multicolumn{1}{c|}{\textbf{63.6}} & \multicolumn{1}{c|}{\textbf{42.4}} & \multicolumn{1}{c|}{\textbf{32.2}} & \multicolumn{1}{c|}{23.6} & \multicolumn{1}{c|}{18.9}\\ \hline
\end{tabular}
} 
\caption{Effect of the network initialization strategy in the PASCAL3D+ for the \textit{Single-path} architecture.}
\label{table:initialization}
\end{table}

Table \ref{table:initialization} shows the main results using the described initialization strategies. In terms of object detection precision, \ie mAP, the initialization of our model, using the PASCAL VOC 2012 datasets is the best option, by a considerable margin, with respect to the ImageNet based strategy. Interestingly, the mAP of our model (63.6) improves the state-of-the-art for the object detection task in the official PASCAL3D+ leaderboard \footnote{Official PASCAL3D+ leaderboard is available at \url{http://cvgl.stanford.edu/projects/pascal3d.html}}, where the best mAP is of 62.5 reported in \cite{MassaBMVC2016}.

In terms of a joint object detection and pose estimation, we also report the mAVP for different sets of views (4, 8, 16 and 24). The ImageNet based initialization reports slightly better results only for the more fine grained setups of 16 and 24 views. When just 4 or 8 views are considered, the initialization process using the PASCAL VOC 2012 is the best option, considering its high detection precision. This first experiment also reveals that it seems to be a trade-off between how good the system is localizing objects and how accurate the pose predictions are. Overall, we conclude that the best initialization strategy is clearly the one based on the PASCAL VOC 2012 dataset. Therefore, for the rest of experiments, we follow this initialization strategy.

\subsubsection{Discrete vs. Continuous approaches analysis}
\label{sec:pascal_loss_analysis}

As we have discussed in Section \ref{sec:loss_pose}, the pose estimation problem can be treated following either a discrete approach, \ie as a classification problem, or a continuous approximation, \ie as a regression problem. One of the main objectives of our study is to shed light on this discussion. 

We have carefully designed all our architectures, so they all can consider a discrete and a continuous approximation to the pose estimation problem. We simply have to \emph{change} the Pose estimation layer, and its associated loss function. Up to four different loss functions are analyzed in these experiments, one for the discrete case and three for the continuous approach.

When the discrete scenario is considered, we follow the cross-entropy loss function in Equation \ref{eq:softmax}. Technically, our architectures consider 360 different classes for the azimuth angle. For each category in the dataset (except for the background), we learn a specific pose estimator, therefore, we need to define a softmax function with a length of $360 \times C$ elements, where $C$ is the number of classes. During learning, we have opted to \emph{mask} the softmax layer, propagating only the error for the elements that correspond to the pose of the foreground class.

For the continuous pose estimation problem, our networks learn to directly perform the regression of the two values corresponding to the conversion to polar coordinates the azimuth angle. We design our deep models to learn a particular regressor for each object category. And again, during learning, only the regressor that corresponds to the associated class label of the sample in the training batch, is allowed to propagate errors. Following this continuous setup, we analyze the three different loss functions introduced in Section \ref{sec:loss_pose}: the Euclidean loss (Eq. \ref{eq:euclideanloss}), the Huber loss (Eq. \ref{eq:huberloss}), and the Cyclic cosine loss (Eq. \ref{eq:cyccosloss}).

\begin{table}[t]
\centering
\resizebox{\columnwidth}{!}{%
\begin{tabular}{lccccc}
\hline
Losses & mAP & mAVP 4 & mAVP 8 & mAVP 16 & mAVP 24 \\ \hline
\\ \hline 
\multicolumn{1}{|l|}{Discrete (Eq. \ref{eq:softmax})}  & \multicolumn{1}{c|}{63.6} & \multicolumn{1}{c|}{42.4} & \multicolumn{1}{c|}{32.2} & \multicolumn{1}{c|}{\textbf{23.6}} & \multicolumn{1}{c|}{\textbf{18.9}}\\ \hline
\multicolumn{1}{|l|}{Euclidean (Eq. \ref{eq:euclideanloss})} & \multicolumn{1}{c|}{64.3} & \multicolumn{1}{c|}{\textbf{47.9}} & \multicolumn{1}{c|}{\textbf{34.7}} & \multicolumn{1}{c|}{23.2} & \multicolumn{1}{c|}{17.6} \\ \hline
\multicolumn{1}{|l|}{Huber (Eq. \ref{eq:huberloss})} & \multicolumn{1}{c|}{\textbf{64.5}} & \multicolumn{1}{c|}{46.1} & \multicolumn{1}{c|}{31.5} & \multicolumn{1}{c|}{20.2} & \multicolumn{1}{c|}{15.2} \\ \hline
\multicolumn{1}{|l|}{Cyclic Cosine (Eq. \ref{eq:cyccosloss})} & \multicolumn{1}{c|}{55.6} & \multicolumn{1}{c|}{42.1} & \multicolumn{1}{c|}{32.2} & \multicolumn{1}{c|}{22.5} & \multicolumn{1}{c|}{17.5} \\ \hline
\end{tabular}
} 
\caption{Loss function analysis for the PASCAL3D+ dataset. Object detection and viewpoint estimation performances are reported.}
\label{tab:losscmp}
\end{table}

Table \ref{tab:losscmp} shows the main results, when the different loss functions are used.
Discrete, Euclidean and Huber losses exhibit a very similar detection performance (mAP). Only when the Cyclic Cosine loss is used, a substantial drop of the detection performance is reported. The reason we find to explain this fact is that during training, the Cyclic Cosine loss can eventually produce larger gradients than the detection loss. This issue causes that the learning process tends to focus more ``attention'' on the pose estimation task, obtaining a deep model with a worse object localization accuracy. A simple adjustment of the $\lambda$ values in Eq. \ref{eq:global_loss_function} did not properly work in our experiments. Another possibility could be to perform a power normalization of the gradients produced by the different losses at the same level of the network. However, we did not explore this option. Instead, we opted for applying the clipping gradient strategy \cite{razvan2013}, with a threshold value of 5. 

If we analyze now the mAVP, where both object detection and viewpoint estimation accuracies are considered, we can observe that, in general, the best performance is reported when the Euclidean loss based model is used. Moreover, within the group of continuous viewpoint estimation models, the Euclidean is the clear winner. Therefore, for the rest of the paper, when a continuous viewpoint model is learned, we use the Euclidean loss. Interestingly, the continuous approach wins the discrete model only when 4 and 8 set of views are considered. For 16 and 24 views, the discrete model retrieves a slightly better performance. In our experiments, we have noted that the continuous pose estimation approaches tend to offer smooth predictions that are concentrated around the most frequent viewpoint of the training set. However, the discrete approach, with a Softmax loss, does not suffer that much from this pose annotation bias.

\begin{figure*}[t]
\centering
\subfloat[Car - Euclidean loss.]{\includegraphics[width=0.45\columnwidth]{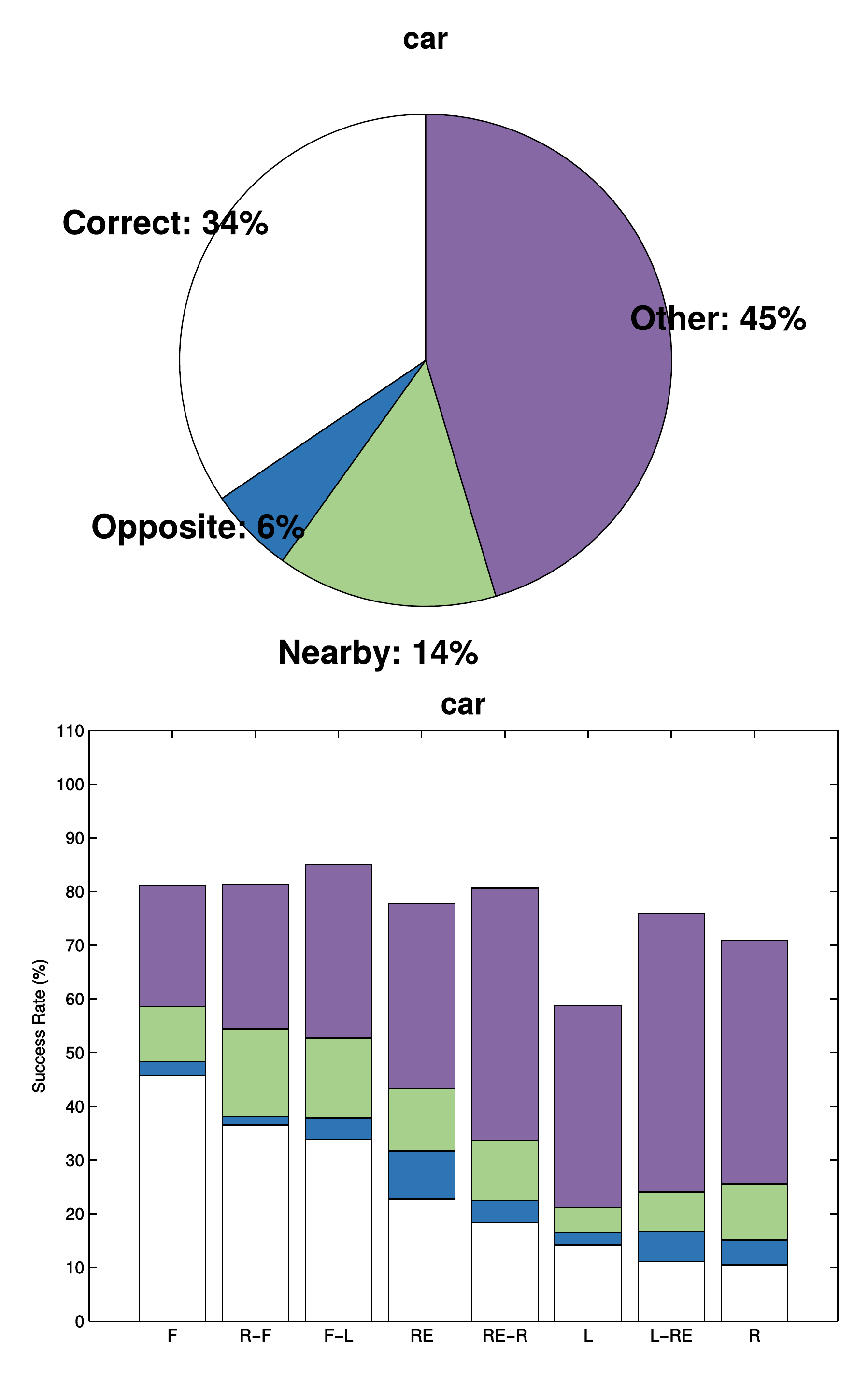} \label{fig:fig:car-conf-euclidean}}  \subfloat[Car - Softmax loss.]{\includegraphics[width=0.45\columnwidth]{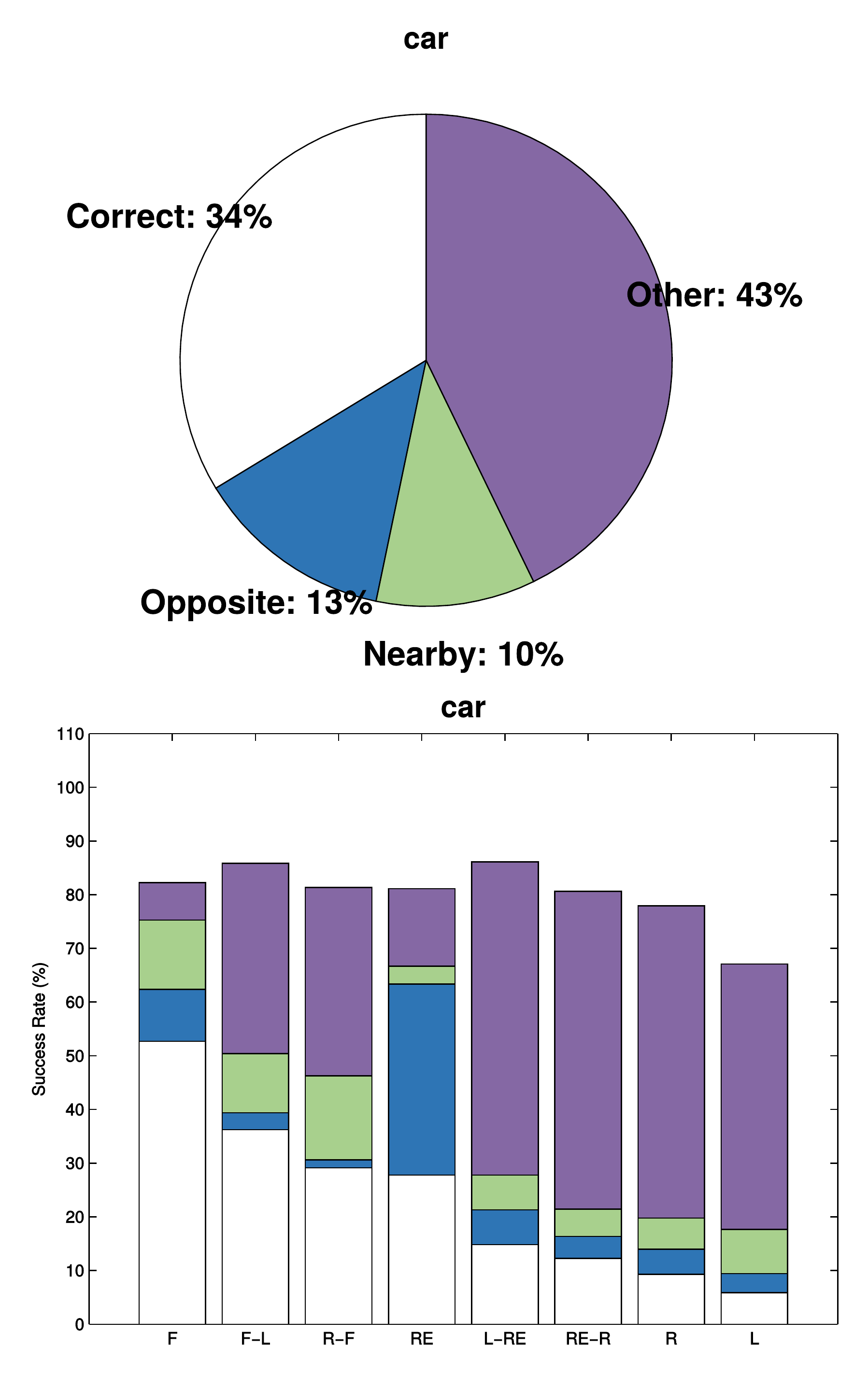} \label{fig:fig:car-conf-discrete}} \subfloat[Bus - Euclidean loss.]{\includegraphics[width=0.45\columnwidth]{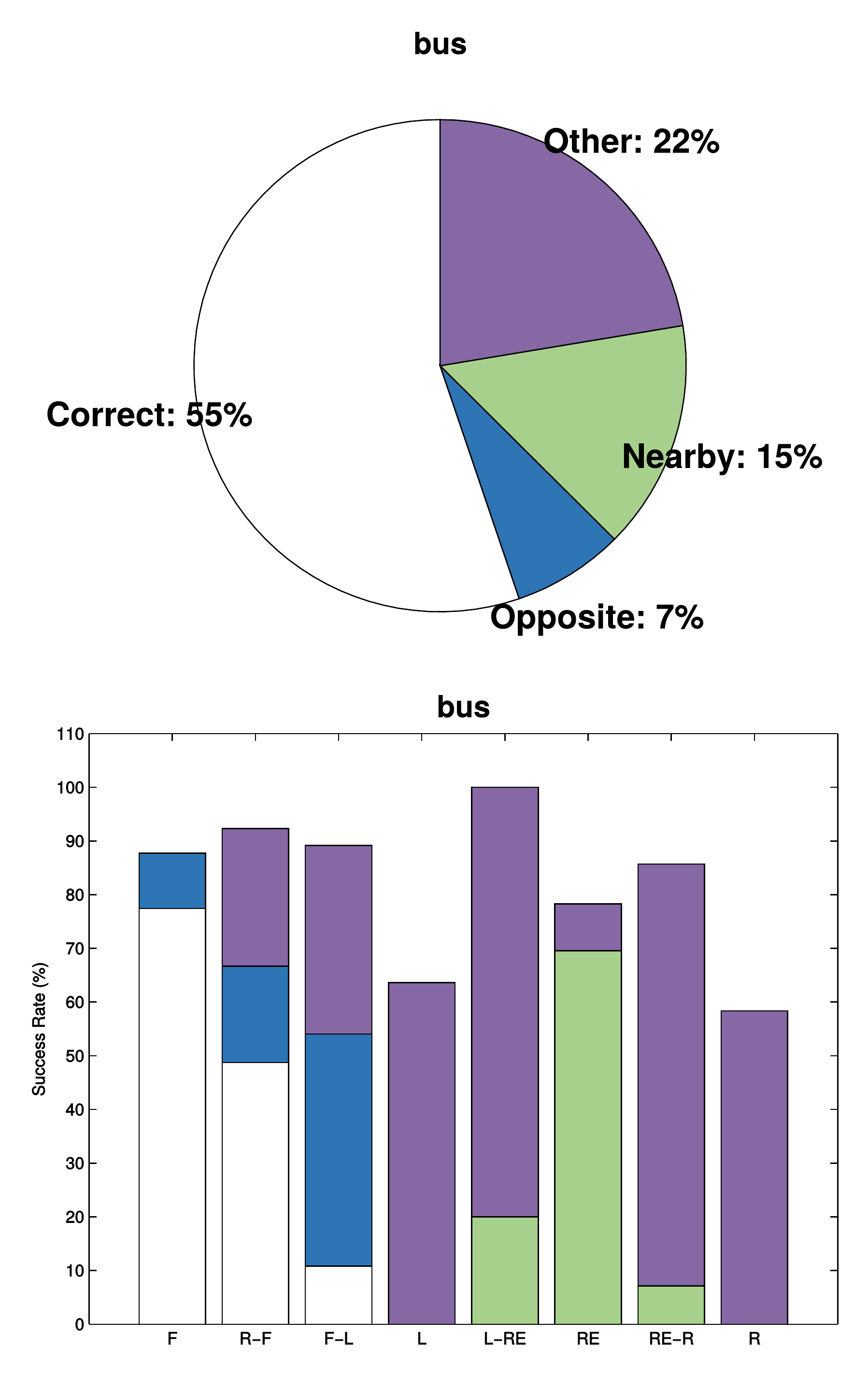} \label{fig:fig:bus-conf-euclidean}}  \subfloat[Bus - Softmax loss.]{\includegraphics[width=0.45\columnwidth]{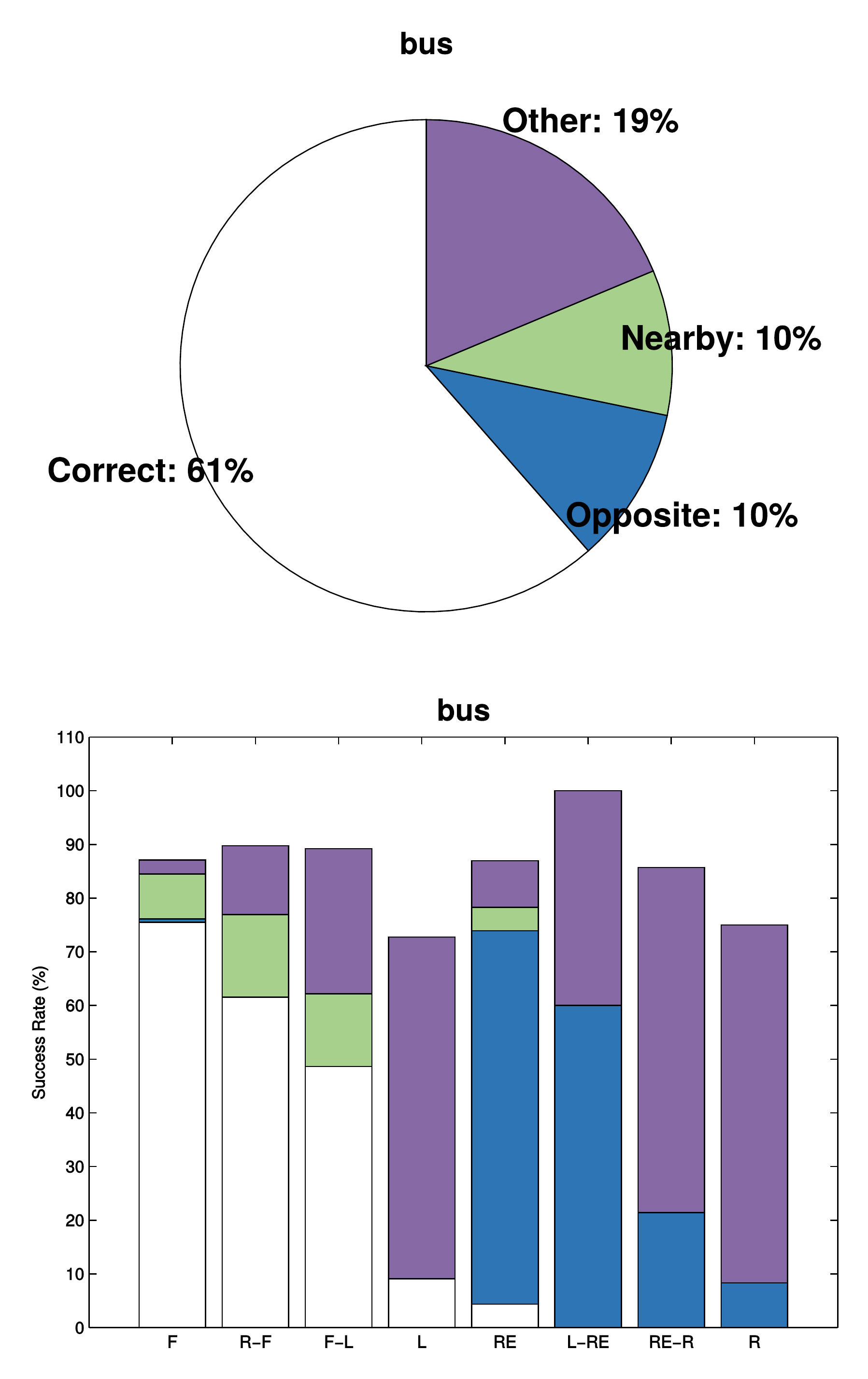} \label{fig:fig:bus-conf-discrete}}
\caption{Viewpoint estimation performance detailed analysis. A comparison between continuous (with Euclidean loss) and discrete (with a Softmax loss) models for categories \emph{Car} and \emph{Bus}. (a) and (b) contain the results for the \emph{car} category, while (c) and (d) show the results for the \emph{bus} class. First row include pie charts showing the general performance of the models, where it is reported the percentage of: correct detections, confusions with opposite viewpoints, confusions with nearby poses, and the rest of errors (Other). Second row shows a detailed analysis, of the same type of errors, considering 8 set of viewpoints (F: Frontal, R-F: Right-Frontal, F-L: Frontal-Left, RE: Rear, RE-R: Rear-Right, L: Left, L-RE: Left-Rear and R: Right).}
\label{fig:discrete_vs_continuous}
\end{figure*}

Figure \ref{fig:discrete_vs_continuous} shows a detailed comparison of the performance between a discrete and a continuous approach for a pair of representative object categories: \emph{car} and \emph{bus}. \emph{Car} is the class with the largest amount of samples in the PASCAL3D+ dataset, \ie 1004 instances of non-difficult objects. The annotated views for cars are distributed quite homogeneously across all the poses, although they are slightly biased towards the frontal and rear views. Category \emph{bus} provides only 301 samples, and the pose is clearly concentrated in the frontal view.

For the category \emph{Car}, Figures \ref{fig:fig:car-conf-euclidean} and \ref{fig:fig:car-conf-discrete} show that the performance of both models (continuous and discrete) are comparable. The continuous pose model tends to get confused with nearby views, while the discrete approach reports more errors with opposite viewpoints. The scenario changes when one inspects the results for the \emph{Bus} object category. Figures \ref{fig:fig:bus-conf-euclidean} and \ref{fig:fig:bus-conf-discrete} show that the performance of the continuous model is slightly worse than the one of the discrete model. Like we detail above, the continuous model tends to concentrate its predictions around the pose annotated bias (\ie the frontal). Observe the bar diagram in \ref{fig:fig:bus-conf-euclidean}, where most of the Rear views are assigned to Frontal views. 

We want to conclude this analysis, adding an additional dimension to the discussion: the influence (in the performance) of the evaluation metric used. The problem of \emph{simultaneous} detection and pose estimation has not been associated with either a clear experimental evaluation process or an evaluation metric. Obviously, part of the problem is that discrete and continuous approaches, being of a different nature, have been evaluated in different ways. As a result, multiple evaluation metrics have been proposed, \eg Pose Estimation Average Precision (PEAP) \cite{Lopez-Sastre2011}, Average Orientation Similarity (AOS) \cite{Geiger2012} and AVP \cite{xiang2014}. We refer the reader to \cite{redondoCabrera2016}, where an extensive analysis of the different evaluation metrics is presented.

\begin{figure}
\centering
\includegraphics[width=0.90\columnwidth]{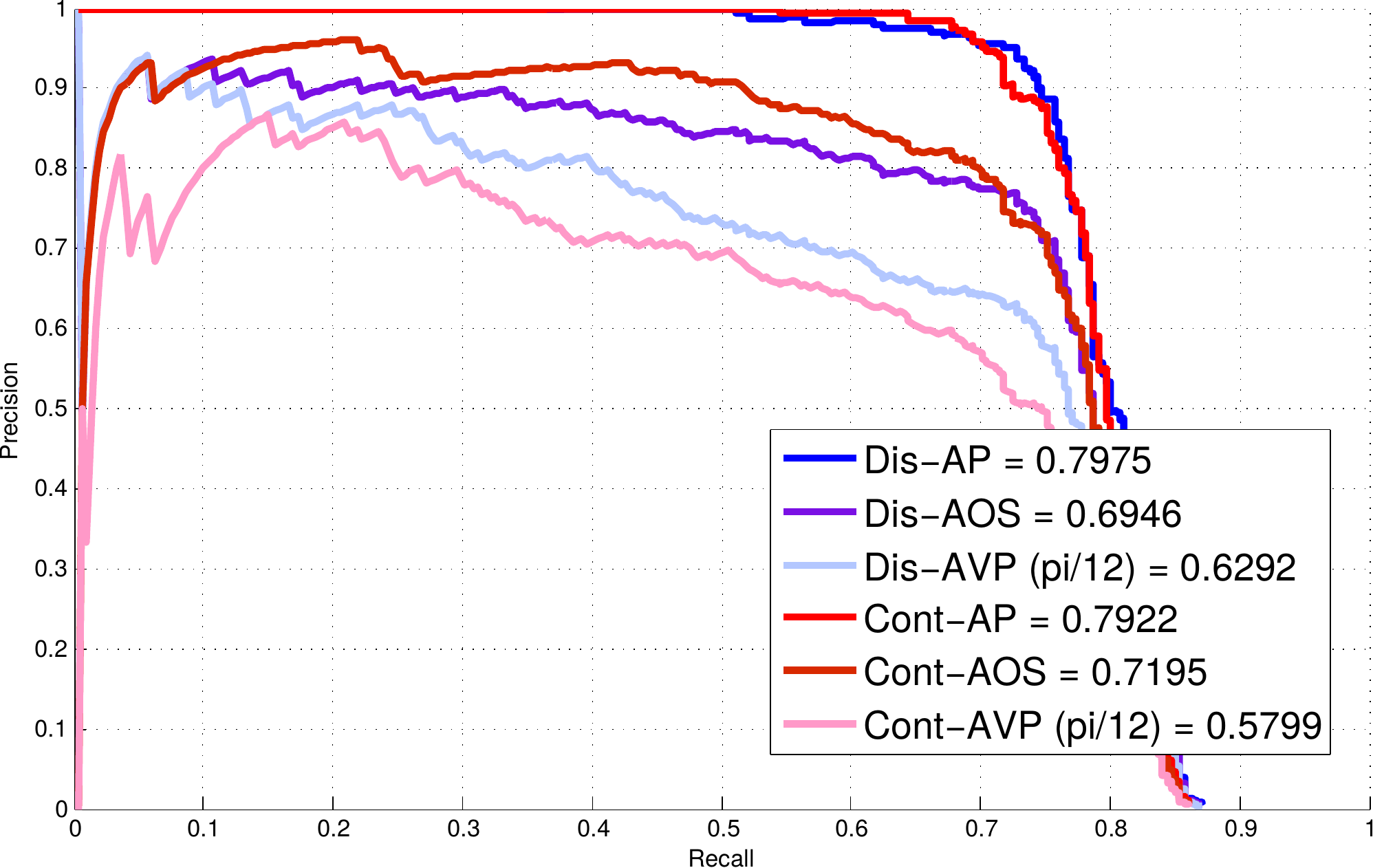}
\caption{Detection and pose estimation performance for the \emph{bus} category. A comparison based on evaluation metrics AOS and AVP, for both continuous (red tonalities) and discrete (blue tonalities) approaches.}
\label{fig:bus_metrics}
\end{figure}

We have compared the performance of the AVP and the AOS metrics. Our experiments reveal that the AVP metric tends to favor discrete approaches, while the AOS metric favors the continuous models. For instance, for the category \emph{bus}, Figure \ref{fig:bus_metrics} shows the precision-recall curves when the different metrics are used. When the AVP metric is used, the discrete approach (Dis-AVP) obtains a higher average precision, compared to the one reported for the continuous model (Cont-AVP). On the other hand, when the AOS metric is followed, the average precision is slightly superior for the continuous model, \ie Dis-AOS $<$ Cont-AOS.

In any case, taking into account the observations made in \cite{redondoCabrera2016}, we would like to remark that the AOS metric is not an adequate measurement of the object detection and
pose estimation problem. In \cite{redondoCabrera2016}, the authors show that this metric is dominated mainly by the detection performance, masquerading the pose estimation precision. Therefore, for the rest of our study, we choose to use an evaluation procedure based only on the AVP metric.

Overall, based on these results, we conclude that continuous viewpoint estimation models tend to accumulate errors at nearby poses, while discrete pose estimation approaches errors are more likely to occur in opposite views. Objectively, errors with close poses are not as important as errors associated with opposite poses. We believe that the continuous models could result more attractive for the problem we are dealing with. However, if the amount of training data is not large enough, and is not well balanced in terms of pose annotations, a discrete estimation model,\ie based on a classifier, is the best option. This is the normal situation in all datasets, and also in the PASCAL3D+. Therefore, \emph{for the rest of our study}, we opt for a discrete model.

\subsubsection{Independent vs Joint object detection and pose estimation}

A quick reading of the scientific literature reveals two main models for tackling the problem of detecting and estimating the pose of object categories. On the one hand we find those who decouple both tasks. The detector is trained and executed separately to locate objects in the images. Subsequently, the pose estimator is responsible for associating a pose to the detected object. On the other hand, we have the models that are trained to solve both tasks together. In this section, we analyze the performance of these two families of works. To do so, we offer a detailed comparison of the proposed architectures in Section \ref{sec:analyzed_models}, with existing state-of-the-art models that belong to one family or another.

We need to start this experimental evaluation making the following observations with respect to the three architecture proposed in this paper. Technically, our 3 network designs present a clear evolution in terms of the degree of coupling of the tasks of detection and pose estimation. Our \textit{Single-path} approach clearly belongs to the \emph{joint} family. Note that in this architecture, all the features of the network are shared for both tasks. With the \textit{Specific-path} architecture we advance one step forward in the decoupling degree. It is a \emph{hybrid} system, where the convolutional layer features are shared, while the FC layers are split into two paths: one for the object localization and one for the pose estimation. Finally we propose the \textit{Specific-network}. Although it should be considered as an architecture belonging to the group of \emph{independent}, we cannot forget that it actually proposes a new paradigm, where both networks, specialized in different tasks, can be trained end-to-end. Note that although the networks learn their characteristics in a decoupled way, the ROIs produced by the network in charge of the location are shared with the network for the estimation of the pose, which somehow conditions their learning. This end-to-end methodology clearly differs from the rest of state-of-the-art \emph{independent} models (\eg \cite{Tulsiani_vpsKps_2015} ).

\begin{figure*}
\centering
\includegraphics[width=0.95\linewidth]{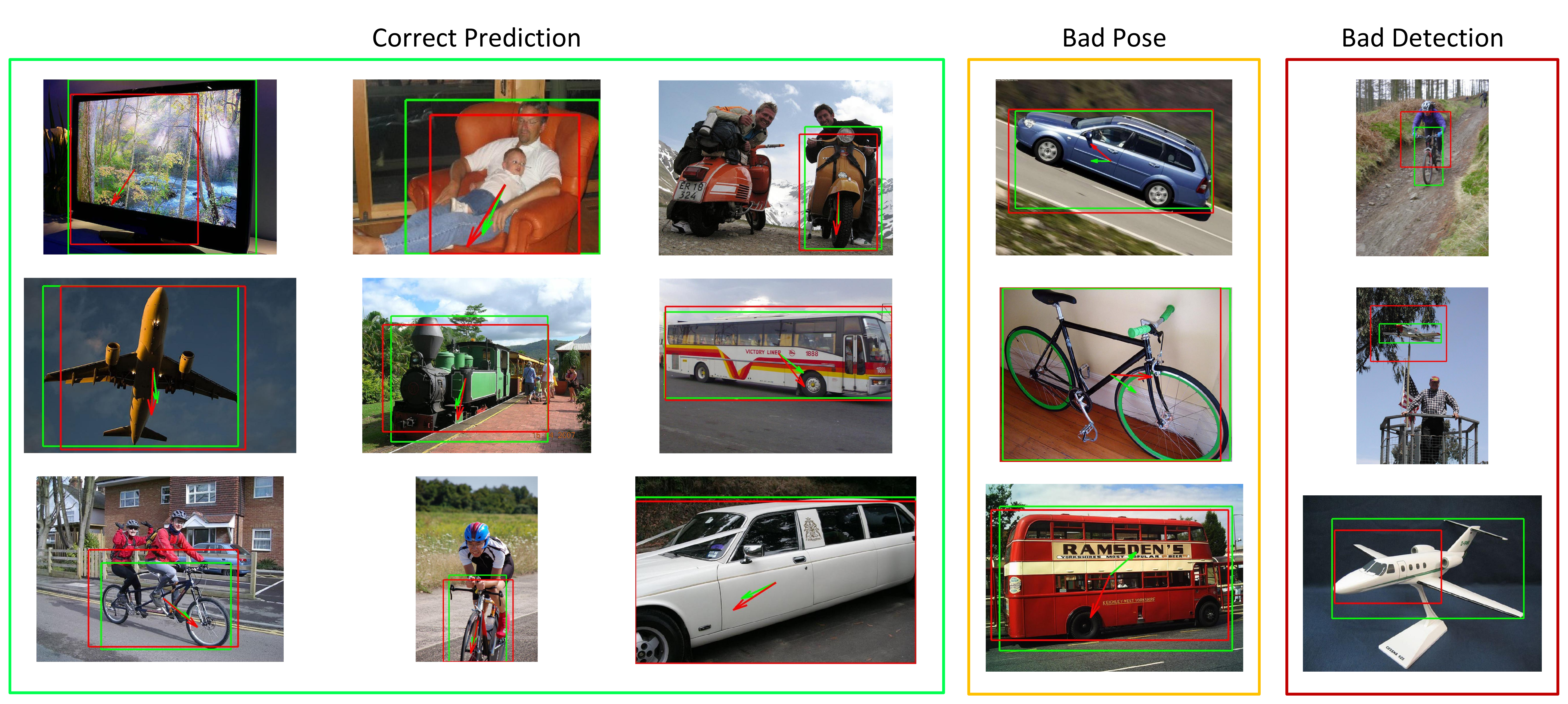} 
\caption{Qualitative results produced by the \textit{Specific-network} in the PASCAL3D+ dataset. In green, we depict the ground truth annotations, while in red we show the results produced by our model. Rectangles correspond to the bounding boxes, while the arrows depict annotated orientations of the objects.}
\label{fig:pascal3d_qualitative}
\end{figure*}

Table \ref{tab:archcmp_pascal3d} shows the results for all of our architectures in the PASCAL3D+ dataset. Overall, our two \emph{independent} models report a better performance than the \textit{Single-path} architecture. For the specific case of 4 set of views, the best performance is given by the \textit{Specific-path} model, which achieves the best AVP for 6 of 11 categories. For the rest of set of views (8, 16 and 24), the best performance is consistently achieved by our \textit{Specific-network} architecture. The obtained results show that the \textit{independent} approaches perform better than \textit{joint} approaches. In Figure \ref{fig:pascal3d_qualitative}, we show some qualitative examples produced by our \textit{Specific-network} architecture. 

\begin{table*}[t]
\centering
\resizebox{0.8\linewidth}{!}{%
\begin{tabular}{lcccccccccccc}
\hline 
\multicolumn{1}{l}{Methods} & \multicolumn{1}{c}{Aero} & \multicolumn{1}{c}{Bike} & \multicolumn{1}{c}{Boat} & \multicolumn{1}{c}{Bus} & \multicolumn{1}{c}{Car} & \multicolumn{1}{c}{Chair} & \multicolumn{1}{c}{Table} & \multicolumn{1}{c}{MBike} & \multicolumn{1}{c}{Sofa} & \multicolumn{1}{c}{Train} & \multicolumn{1}{c}{Monitor} & \multicolumn{1}{c}{Avg.} \\ \hline 
\multicolumn{13}{c}{AP Object Detection}                                                                                                                                                                                                                                                                                                              \\ \hline
\rowcolor{tableGray}
\multicolumn{1}{|l|}{Single-path}         & \multicolumn{1}{c|}{78.1}     & \multicolumn{1}{c|}{\textbf{74.3}}     & \multicolumn{1}{c|}{47.2}    & \multicolumn{1}{c|}{\textbf{79.7}}    & \multicolumn{1}{c|}{70.2}      & \multicolumn{1}{c|}{28.2}      & \multicolumn{1}{c|}{\textbf{53.0}}          & \multicolumn{1}{c|}{76.0}     & \multicolumn{1}{c|}{52.0}      & \multicolumn{1}{c|}{\textbf{79.5}}        & \multicolumn{1}{c|}{60.8}    & \multicolumn{1}{c|}{63.6}        \\ \hline
\multicolumn{1}{|l|}{Specific-path}        & \multicolumn{1}{c|}{\textbf{78.5}}         & \multicolumn{1}{c|}{73.1}     & \multicolumn{1}{c|}{\textbf{49.3}}     & \multicolumn{1}{c|}{79.2}    & \multicolumn{1}{c|}{\textbf{70.3}}    & \multicolumn{1}{c|}{\textbf{32.3}}      & \multicolumn{1}{c|}{52.7}      & \multicolumn{1}{c|}{78.0}          & \multicolumn{1}{c|}{\textbf{58.0}}     & \multicolumn{1}{c|}{77.9}      & \multicolumn{1}{c|}{\textbf{64.6}}        & \multicolumn{1}{c|}{\textbf{64.9}}    \\ \hline
\multicolumn{1}{|l|}{Specific-network}        & \multicolumn{1}{c|}{77.8}         & \multicolumn{1}{c|}{74.2}     & \multicolumn{1}{c|}{47.9}     & \multicolumn{1}{c|}{78.7}    & \multicolumn{1}{c|}{\textbf{70.3}}    & \multicolumn{1}{c|}{30.7}      & \multicolumn{1}{c|}{52.9}      & \multicolumn{1}{c|}{\textbf{78.1}}          & \multicolumn{1}{c|}{56.5}     & \multicolumn{1}{c|}{77.7}      & \multicolumn{1}{c|}{62.7}        & \multicolumn{1}{c|}{64.3}    \\ \hline 
\multicolumn{13}{c}{AVP 4 Views - Joint Object Detection and Pose Estimation}                                                                                                                                                                                                                                                                                                              \\ \hline
\rowcolor{tableGray}
\multicolumn{1}{|l|}{Single-path}         & \multicolumn{1}{c|}{52.4}     & \multicolumn{1}{c|}{41.7}     & \multicolumn{1}{c|}{18.6}    & \multicolumn{1}{c|}{66.2}    & \multicolumn{1}{c|}{45.3}      & \multicolumn{1}{c|}{14.2}      & \multicolumn{1}{c|}{26.1}          & \multicolumn{1}{c|}{44.7}     & \multicolumn{1}{c|}{40.4}      & \multicolumn{1}{c|}{63.7}        & \multicolumn{1}{c|}{52.9}    & \multicolumn{1}{c|}{42.4}        \\ \hline
\multicolumn{1}{|l|}{Specific-path}        & \multicolumn{1}{c|}{56.7}         & \multicolumn{1}{c|}{54.7}     & \multicolumn{1}{c|}{\textbf{24.1}}     & \multicolumn{1}{c|}{66.2}    & \multicolumn{1}{c|}{50.2}    & \multicolumn{1}{c|}{\textbf{17.3}}      & \multicolumn{1}{c|}{\textbf{30.1}}      & \multicolumn{1}{c|}{55.7}          & \multicolumn{1}{c|}{\textbf{44.0}}     & \multicolumn{1}{c|}{\textbf{61.6}}      & \multicolumn{1}{c|}{\textbf{60.4}}        & \multicolumn{1}{c|}{\textbf{47.4}}    \\ \hline
\multicolumn{1}{|l|}{Specific-network}        & \multicolumn{1}{c|}{\textbf{58.4}}         & \multicolumn{1}{c|}{\textbf{57.0}}     & \multicolumn{1}{c|}{23.2}     & \multicolumn{1}{c|}{\textbf{66.3}}    & \multicolumn{1}{c|}{\textbf{53.3}}    & \multicolumn{1}{c|}{16.9}      & \multicolumn{1}{c|}{27.9}      & \multicolumn{1}{c|}{\textbf{60.9}}          & \multicolumn{1}{c|}{41.5}     & \multicolumn{1}{c|}{60.1}      & \multicolumn{1}{c|}{52.6}        & \multicolumn{1}{c|}{47.1}    \\ \hline 
\multicolumn{13}{c}{AVP 8 Views - Joint Object Detection and Pose Estimation}                                                                                                                                                                                                                                                                                                             \\ \hline
\rowcolor{tableGray}
\multicolumn{1}{|l|}{Single-path}        & \multicolumn{1}{c|}{42.9}         & \multicolumn{1}{c|}{28.9}     & \multicolumn{1}{c|}{11.1}     & \multicolumn{1}{c|}{52.7}    & \multicolumn{1}{c|}{38.8}    & \multicolumn{1}{c|}{10.5}      & \multicolumn{1}{c|}{18.1}      & \multicolumn{1}{c|}{32.0}          & \multicolumn{1}{c|}{28.3}     & \multicolumn{1}{c|}{50.2}      & \multicolumn{1}{c|}{40.5}        & \multicolumn{1}{c|}{32.2}    \\ \hline
\multicolumn{1}{|l|}{Specific-path}        & \multicolumn{1}{c|}{47.2}         & \multicolumn{1}{c|}{38.3}     & \multicolumn{1}{c|}{\textbf{16.3}}     & \multicolumn{1}{c|}{47.2}    & \multicolumn{1}{c|}{43.0}    & \multicolumn{1}{c|}{12.8}      & \multicolumn{1}{c|}{\textbf{25.5}}      & \multicolumn{1}{c|}{47.5}          & \multicolumn{1}{c|}{33.2}     & \multicolumn{1}{c|}{\textbf{53.4}}      & \multicolumn{1}{c|}{\textbf{43.5}}        & \multicolumn{1}{c|}{37.1}    \\ \hline
\multicolumn{1}{|l|}{Specific-network}        & \multicolumn{1}{c|}{\textbf{51.3}}         & \multicolumn{1}{c|}{\textbf{43.2}}     & \multicolumn{1}{c|}{14.4}     & \multicolumn{1}{c|}{\textbf{54.6}}    & \multicolumn{1}{c|}{\textbf{46.1}}    & \multicolumn{1}{c|}{\textbf{13.3}}      & \multicolumn{1}{c|}{21.8}      & \multicolumn{1}{c|}{\textbf{48.4}}          & \multicolumn{1}{c|}{\textbf{33.8}}     & \multicolumn{1}{c|}{49.4}      & \multicolumn{1}{c|}{41.7}        & \multicolumn{1}{c|}{\textbf{38.2}}    \\ \hline 
\multicolumn{13}{c}{AVP 16 Views - Joint Object Detection and Pose Estimation}                                                                                                                                                                                                                                                                                                              \\ \hline
\rowcolor{tableGray}
\multicolumn{1}{|l|}{Single-path}        & \multicolumn{1}{c|}{22.8}         & \multicolumn{1}{c|}{19.5}     & \multicolumn{1}{c|}{7.8}     & \multicolumn{1}{c|}{54.4}    & \multicolumn{1}{c|}{31.8}    & \multicolumn{1}{c|}{6.8}      & \multicolumn{1}{c|}{14.0}      & \multicolumn{1}{c|}{20.5}          & \multicolumn{1}{c|}{15.6}     & \multicolumn{1}{c|}{\textbf{42.9}}      & \multicolumn{1}{c|}{23.6}        & \multicolumn{1}{c|}{23.6}    \\ \hline
\multicolumn{1}{|l|}{Specific-path}        & \multicolumn{1}{c|}{33.4} & \multicolumn{1}{c|}{25.9}     & \multicolumn{1}{c|}{10.1}    & \multicolumn{1}{c|}{51.3}    & \multicolumn{1}{c|}{32.7}      & \multicolumn{1}{c|}{8.0}      & \multicolumn{1}{c|}{20.1}          & \multicolumn{1}{c|}{23.8}     & \multicolumn{1}{c|}{\textbf{25.9}}      & \multicolumn{1}{c|}{38.0}        & \multicolumn{1}{c|}{\textbf{32.5}}         & \multicolumn{1}{c|}{27.4}     \\ \hline
\multicolumn{1}{|l|}{Specific-network}        & \multicolumn{1}{c|}{\textbf{36.7}}         & \multicolumn{1}{c|}{\textbf{30.5}}     & \multicolumn{1}{c|}{\textbf{11.7}}     & \multicolumn{1}{c|}{\textbf{57.4}}    & \multicolumn{1}{c|}{\textbf{39.7}}    & \multicolumn{1}{c|}{\textbf{8.9}}      & \multicolumn{1}{c|}{\textbf{21.8}}      & \multicolumn{1}{c|}{\textbf{29.6}}          & \multicolumn{1}{c|}{25.5}     & \multicolumn{1}{c|}{38.0}      & \multicolumn{1}{c|}{31.9}        & \multicolumn{1}{c|}{\textbf{30.2}}    \\ \hline 
\multicolumn{13}{c}{AVP 24 Views - Joint Object Detection and Pose Estimation}                                                                                                                                                                                                                                                                                                              \\ \hline
\rowcolor{tableGray}
\multicolumn{1}{|l|}{Single-path}        & \multicolumn{1}{c|}{18.1}         & \multicolumn{1}{c|}{15.3}     & \multicolumn{1}{c|}{4.4}     & \multicolumn{1}{c|}{44.8}    & \multicolumn{1}{c|}{27.2}    & \multicolumn{1}{c|}{5.2}      & \multicolumn{1}{c|}{11.8}      & \multicolumn{1}{c|}{13.7}          & \multicolumn{1}{c|}{14.0}     & \multicolumn{1}{c|}{\textbf{36.9}}      & \multicolumn{1}{c|}{16.9}        & \multicolumn{1}{c|}{18.9}    \\ \hline
\multicolumn{1}{|l|}{Specific-path}        & \multicolumn{1}{c|}{\textbf{26.0}}         & \multicolumn{1}{c|}{18.3}     & \multicolumn{1}{c|}{7.7}     & \multicolumn{1}{c|}{40.6}    & \multicolumn{1}{c|}{29.3}    & \multicolumn{1}{c|}{5.2}      & \multicolumn{1}{c|}{15.9}      & \multicolumn{1}{c|}{18.4}          & \multicolumn{1}{c|}{\textbf{20.3}}     & \multicolumn{1}{c|}{36.7}      & \multicolumn{1}{c|}{\textbf{24.4}}        & \multicolumn{1}{c|}{22.1}    \\ \hline
\multicolumn{1}{|l|}{Specific-network}        & \multicolumn{1}{c|}{22.9}         & \multicolumn{1}{c|}{\textbf{21.8}}     & \multicolumn{1}{c|}{\textbf{8.8}}     & \multicolumn{1}{c|}{\textbf{45.0}}    & \multicolumn{1}{c|}{\textbf{33.2}}    & \multicolumn{1}{c|}{\textbf{7.0}}      & \multicolumn{1}{c|}{\textbf{18.2}}      & \multicolumn{1}{c|}{\textbf{20.8}}          & \multicolumn{1}{c|}{16.9}     & \multicolumn{1}{c|}{33.4}      & \multicolumn{1}{c|}{21.8}        & \multicolumn{1}{c|}{\textbf{22.7}}    \\ \hline 
\end{tabular}
} 
\caption{Object detection and pose estimation results in the PASCAL3D+ dataset. Comparison between all our architectures. In gray color we show our \emph{joint} solution, \ie the \textit{Single-path} architecture. The remaining architectures (\textit{Specific-path} and \textit{Specific-network}) can be classified in the group of \textit{independent} approaches.}
\label{tab:archcmp_pascal3d}
\end{table*}

\begin{table*}[t]
\centering
\resizebox{0.8\linewidth}{!}{%
\begin{tabular}{lccccccccccccc}
\hline 
\multicolumn{1}{l}{Methods} & \multicolumn{1}{c}{Aero} & \multicolumn{1}{c}{Bike} & \multicolumn{1}{c}{Boat} & \multicolumn{1}{c}{Bus} & \multicolumn{1}{c}{Car} & \multicolumn{1}{c}{Chair} & \multicolumn{1}{c}{Table} & \multicolumn{1}{c}{MBike} & \multicolumn{1}{c}{Sofa} & \multicolumn{1}{c}{Train} & \multicolumn{1}{c}{Monitor} & \multicolumn{1}{c}{mAVP} & \multicolumn{1}{c}{mAP} \\ \hline 
\multicolumn{14}{c}{AVP 4 Views - Joint Object Detection and Pose Estimation}                                                                                                                                                                                                                                                                                                              \\ \hline
\rowcolor{tableGray}
\multicolumn{1}{|l|}{VDPM \cite{xiang2014}}        & \multicolumn{1}{c|}{34.6}         & \multicolumn{1}{c|}{41.7}     & \multicolumn{1}{c|}{1.5}     & \multicolumn{1}{c|}{26.1}    & \multicolumn{1}{c|}{20.2}    & \multicolumn{1}{c|}{6.8}      & \multicolumn{1}{c|}{3.1}      & \multicolumn{1}{c|}{30.4}          & \multicolumn{1}{c|}{5.1}     & \multicolumn{1}{c|}{10.7}      & \multicolumn{1}{c|}{34.7}        & \multicolumn{1}{c|}{19.5} & \multicolumn{1}{c|}{26.8}     \\ \hline
\rowcolor{tableGray}
\multicolumn{1}{|l|}{DPM-VOC+VP \cite{Pepik2012}}        & \multicolumn{1}{c|}{37.4}         & \multicolumn{1}{c|}{43.9}     & \multicolumn{1}{c|}{0.3}     & \multicolumn{1}{c|}{48.6}    & \multicolumn{1}{c|}{36.9}    & \multicolumn{1}{c|}{6.1}      & \multicolumn{1}{c|}{2.1}      & \multicolumn{1}{c|}{31.8}          & \multicolumn{1}{c|}{11.8}     & \multicolumn{1}{c|}{11.1}      & \multicolumn{1}{c|}{32.2}        & \multicolumn{1}{c|}{23.8}  & \multicolumn{1}{c|}{27.0}   \\ \hline
\rowcolor{tableGray}
\multicolumn{1}{|l|}{Craft-CNN \cite{MassaBMVC2016}}        & \multicolumn{1}{c|}{-}         & \multicolumn{1}{c|}{-}     & \multicolumn{1}{c|}{-}     & \multicolumn{1}{c|}{-}    & \multicolumn{1}{c|}{-}    & \multicolumn{1}{c|}{-}      & \multicolumn{1}{c|}{-}      & \multicolumn{1}{c|}{-}          & \multicolumn{1}{c|}{-}     & \multicolumn{1}{c|}{-}      & \multicolumn{1}{c|}{-}        & \multicolumn{1}{c|}{-} & \multicolumn{1}{c|}{-}    \\ \hline
\rowcolor{tableGray}
\multicolumn{1}{|l|}{Single-Shot \cite{poirson2016_singleshot}}        & \multicolumn{1}{c|}{64.6}         & \multicolumn{1}{c|}{62.1}     & \multicolumn{1}{c|}{26.8-}     & \multicolumn{1}{c|}{70.0}    & \multicolumn{1}{c|}{51.4}    & \multicolumn{1}{c|}{11.3}      & \multicolumn{1}{c|}{40.7}      & \multicolumn{1}{c|}{62.7}          & \multicolumn{1}{c|}{40.6}     & \multicolumn{1}{c|}{65.9}      & \multicolumn{1}{c|}{61.3}        & \multicolumn{1}{c|}{\textbf{50.7}}  & \multicolumn{1}{c|}{61.0}  \\ \hline
\rowcolor{tableGray}
\multicolumn{1}{|l|}{SubCNN \cite{xiang2016subcategory}}        & \multicolumn{1}{c|}{61.4}         & \multicolumn{1}{c|}{60.4}     & \multicolumn{1}{c|}{21.1}     & \multicolumn{1}{c|}{63.0}    & \multicolumn{1}{c|}{48.7}    & \multicolumn{1}{c|}{23.8}      & \multicolumn{1}{c|}{17.4}      & \multicolumn{1}{c|}{60.7}        & \multicolumn{1}{c|}{47.8}     & \multicolumn{1}{c|}{55.9}      & \multicolumn{1}{c|}{62.3}        & \multicolumn{1}{c|}{47.5} & \multicolumn{1}{c|}{60.7} \\ \hline
\rowcolor{tableGray}
\multicolumn{1}{|l|}{RenderCNN \cite{Su_2015_ICCV}}        & \multicolumn{1}{c|}{50.0}         & \multicolumn{1}{c|}{50.5}     & \multicolumn{1}{c|}{15.1}     & \multicolumn{1}{c|}{57.1}    & \multicolumn{1}{c|}{41.8}    & \multicolumn{1}{c|}{15.7}      & \multicolumn{1}{c|}{18.6}      & \multicolumn{1}{c|}{50.8}          & \multicolumn{1}{c|}{28.4}     & \multicolumn{1}{c|}{46.1}      & \multicolumn{1}{c|}{58.2}        & \multicolumn{1}{c|}{39.7} & \multicolumn{1}{c|}{56.9}   \\ \hline
\multicolumn{1}{|l|}{VP\&KP \cite{Tulsiani_vpsKps_2015}}        & \multicolumn{1}{c|}{63.1}         & \multicolumn{1}{c|}{59.4}     & \multicolumn{1}{c|}{20.3}     & \multicolumn{1}{c|}{69.8}    & \multicolumn{1}{c|}{55.2}    & \multicolumn{1}{c|}{25.1}      & \multicolumn{1}{c|}{24.3}      & \multicolumn{1}{c|}{61.1}          & \multicolumn{1}{c|}{43.8}     & \multicolumn{1}{c|}{59.4}      & \multicolumn{1}{c|}{55.4}        & \multicolumn{1}{c|}{49.1}  & \multicolumn{1}{c|}{56.9}   \\ \hline
\multicolumn{1}{|l|}{Specific-network}        & \multicolumn{1}{c|}{58.4}         & \multicolumn{1}{c|}{57.0}     & \multicolumn{1}{c|}{23.2}     & \multicolumn{1}{c|}{66.3}    & \multicolumn{1}{c|}{53.3}    & \multicolumn{1}{c|}{16.9}      & \multicolumn{1}{c|}{27.9}      & \multicolumn{1}{c|}{60.9}          & \multicolumn{1}{c|}{41.5}     & \multicolumn{1}{c|}{60.1}      & \multicolumn{1}{c|}{52.6}        & \multicolumn{1}{c|}{47.1} & \multicolumn{1}{c|}{\textbf{64.3}}   \\ \hline 
\multicolumn{14}{c}{AVP 8 Views - Joint Object Detection and Pose Estimation}                                                                                                                                                                                                                                                                                                             \\ \hline
\rowcolor{tableGray}
\multicolumn{1}{|l|}{VDPM \cite{xiang2014}}	& \multicolumn{1}{c|}{23.4}         & \multicolumn{1}{c|}{36.5}     & \multicolumn{1}{c|}{1.0}     & \multicolumn{1}{c|}{35.5}    & \multicolumn{1}{c|}{23.5}    & \multicolumn{1}{c|}{5.8}      & \multicolumn{1}{c|}{3.6}      & \multicolumn{1}{c|}{25.1}          & \multicolumn{1}{c|}{12.5}     & \multicolumn{1}{c|}{10.9}      & \multicolumn{1}{c|}{27.4}        & \multicolumn{1}{c|}{18.7} & \multicolumn{1}{c|}{29.9} \\ \hline
\rowcolor{tableGray}
\multicolumn{1}{|l|}{DPM-VOC+VP \cite{Pepik2012}}        & \multicolumn{1}{c|}{28.6}         & \multicolumn{1}{c|}{40.3}     & \multicolumn{1}{c|}{0.2}     & \multicolumn{1}{c|}{38.0}    & \multicolumn{1}{c|}{36.6}    & \multicolumn{1}{c|}{9.4}      & \multicolumn{1}{c|}{2.6}      & \multicolumn{1}{c|}{32.0}          & \multicolumn{1}{c|}{11.0}     & \multicolumn{1}{c|}{9.8}      & \multicolumn{1}{c|}{28.6}        & \multicolumn{1}{c|}{21.5}  & \multicolumn{1}{c|}{28.3}   \\ \hline
\rowcolor{tableGray}
\multicolumn{1}{|l|}{Craft-CNN \cite{MassaBMVC2016}}        & \multicolumn{1}{c|}{-}         & \multicolumn{1}{c|}{-}     & \multicolumn{1}{c|}{-}     & \multicolumn{1}{c|}{-}    & \multicolumn{1}{c|}{-}    & \multicolumn{1}{c|}{-}      & \multicolumn{1}{c|}{-}      & \multicolumn{1}{c|}{-}          & \multicolumn{1}{c|}{-}     & \multicolumn{1}{c|}{-}      & \multicolumn{1}{c|}{-}        & \multicolumn{1}{c|}{-} & \multicolumn{1}{c|}{-}   \\ \hline
\rowcolor{tableGray}
\multicolumn{1}{|l|}{Single-Shot \cite{poirson2016_singleshot}}        & \multicolumn{1}{c|}{58.7}         & \multicolumn{1}{c|}{56.4}     & \multicolumn{1}{c|}{19.9}     & \multicolumn{1}{c|}{62.4}    & \multicolumn{1}{c|}{42.2}    & \multicolumn{1}{c|}{10.6}      & \multicolumn{1}{c|}{34.7}      & \multicolumn{1}{c|}{58.6}          & \multicolumn{1}{c|}{38.8}     & \multicolumn{1}{c|}{61.2}      & \multicolumn{1}{c|}{49.7}        & \multicolumn{1}{c|}{\textbf{45.1}} & \multicolumn{1}{c|}{60.4}   \\ \hline
\rowcolor{tableGray}
\multicolumn{1}{|l|}{SubCNN \cite{xiang2016subcategory}}        & \multicolumn{1}{c|}{48.8}         & \multicolumn{1}{c|}{36.3}     & \multicolumn{1}{c|}{16.4}     & \multicolumn{1}{c|}{39.8}    & \multicolumn{1}{c|}{37.2}    & \multicolumn{1}{c|}{19.1}      & \multicolumn{1}{c|}{13.2}      & \multicolumn{1}{c|}{37.0}        & \multicolumn{1}{c|}{32.1}     & \multicolumn{1}{c|}{44.4}      & \multicolumn{1}{c|}{26.9}        & \multicolumn{1}{c|}{31.9} & \multicolumn{1}{c|}{60.7}   \\ \hline
\rowcolor{tableGray}
\multicolumn{1}{|l|}{RenderCNN \cite{Su_2015_ICCV}}        & \multicolumn{1}{c|}{44.5}         & \multicolumn{1}{c|}{41.1}     & \multicolumn{1}{c|}{10.1}     & \multicolumn{1}{c|}{48.0}    & \multicolumn{1}{c|}{36.6}    & \multicolumn{1}{c|}{13.7}      & \multicolumn{1}{c|}{15.1}      & \multicolumn{1}{c|}{39.9}          & \multicolumn{1}{c|}{26.8}     & \multicolumn{1}{c|}{39.1}      & \multicolumn{1}{c|}{46.5}        & \multicolumn{1}{c|}{32.9} & \multicolumn{1}{c|}{56.9}   \\ \hline
\multicolumn{1}{|l|}{VP\&KP \cite{Tulsiani_vpsKps_2015}}        & \multicolumn{1}{c|}{57.5}         & \multicolumn{1}{c|}{54.8}     & \multicolumn{1}{c|}{18.9}     & \multicolumn{1}{c|}{59.4}    & \multicolumn{1}{c|}{51.5}    & \multicolumn{1}{c|}{24.7}      & \multicolumn{1}{c|}{20.5}      & \multicolumn{1}{c|}{59.5}          & \multicolumn{1}{c|}{43.7}     & \multicolumn{1}{c|}{53.3}      & \multicolumn{1}{c|}{45.6}        & \multicolumn{1}{c|}{44.5}  & \multicolumn{1}{c|}{56.9}   \\ \hline
\multicolumn{1}{|l|}{Specific-network}        & \multicolumn{1}{c|}{51.3}         & \multicolumn{1}{c|}{43.2}     & \multicolumn{1}{c|}{14.4}     & \multicolumn{1}{c|}{54.6}    & \multicolumn{1}{c|}{46.1}    & \multicolumn{1}{c|}{13.3}      & \multicolumn{1}{c|}{21.8}      & \multicolumn{1}{c|}{48.4}          & \multicolumn{1}{c|}{33.8}     & \multicolumn{1}{c|}{49.4}      & \multicolumn{1}{c|}{41.7}        & \multicolumn{1}{c|}{38.2} & \multicolumn{1}{c|}{\textbf{64.3}}   \\ \hline 
\multicolumn{14}{c}{AVP 16 Views - Joint Object Detection and Pose Estimation}                                                                                                                                                                                                                                                                                                              \\ \hline
\rowcolor{tableGray}
\multicolumn{1}{|l|}{VDPM \cite{xiang2014}}       & \multicolumn{1}{c|}{15.4}         & \multicolumn{1}{c|}{18.4}     & \multicolumn{1}{c|}{0.5}     & \multicolumn{1}{c|}{46.9}    & \multicolumn{1}{c|}{18.1}    & \multicolumn{1}{c|}{6.0}      & \multicolumn{1}{c|}{2.2}      & \multicolumn{1}{c|}{16.1}          & \multicolumn{1}{c|}{10.0}     & \multicolumn{1}{c|}{22.1}      & \multicolumn{1}{c|}{16.3}        & \multicolumn{1}{c|}{15.6} & \multicolumn{1}{c|}{30.0}   \\ \hline
\rowcolor{tableGray}
\multicolumn{1}{|l|}{DPM-VOC+VP \cite{Pepik2012}}        & \multicolumn{1}{c|}{15.9}         & \multicolumn{1}{c|}{22.9}     & \multicolumn{1}{c|}{0.3}     & \multicolumn{1}{c|}{49.0}    & \multicolumn{1}{c|}{29.6}    & \multicolumn{1}{c|}{6.1}      & \multicolumn{1}{c|}{2.3}      & \multicolumn{1}{c|}{16.7}          & \multicolumn{1}{c|}{7.1}     & \multicolumn{1}{c|}{20.2}      & \multicolumn{1}{c|}{19.9}        & \multicolumn{1}{c|}{17.3}  & \multicolumn{1}{c|}{28.3}   \\ \hline
\rowcolor{tableGray}
\multicolumn{1}{|l|}{Craft-CNN \cite{MassaBMVC2016}}        & \multicolumn{1}{c|}{-}         & \multicolumn{1}{c|}{-}     & \multicolumn{1}{c|}{-}     & \multicolumn{1}{c|}{-}    & \multicolumn{1}{c|}{-}    & \multicolumn{1}{c|}{-}      & \multicolumn{1}{c|}{-}      & \multicolumn{1}{c|}{-}          & \multicolumn{1}{c|}{-}     & \multicolumn{1}{c|}{-}      & \multicolumn{1}{c|}{-}        & \multicolumn{1}{c|}{-} & \multicolumn{1}{c|}{-}   \\ \hline
\rowcolor{tableGray}
\multicolumn{1}{|l|}{Single-Shot \cite{poirson2016_singleshot}}        & \multicolumn{1}{c|}{46.1}         & \multicolumn{1}{c|}{39.6}     & \multicolumn{1}{c|}{13.6}     & \multicolumn{1}{c|}{56.0}    & \multicolumn{1}{c|}{36.8}    & \multicolumn{1}{c|}{6.4}      & \multicolumn{1}{c|}{23.5}      & \multicolumn{1}{c|}{41.8}          & \multicolumn{1}{c|}{27.0}     & \multicolumn{1}{c|}{38.8}      & \multicolumn{1}{c|}{36.4}        & \multicolumn{1}{c|}{33.3} & \multicolumn{1}{c|}{60.0}   \\ \hline
\rowcolor{tableGray}
\multicolumn{1}{|l|}{SubCNN \cite{xiang2016subcategory}}        & \multicolumn{1}{c|}{28.0}         & \multicolumn{1}{c|}{23.7}     & \multicolumn{1}{c|}{10.0}     & \multicolumn{1}{c|}{50.8}    & \multicolumn{1}{c|}{31.4}    & \multicolumn{1}{c|}{14.3}      & \multicolumn{1}{c|}{9.4}      & \multicolumn{1}{c|}{23.4}          & \multicolumn{1}{c|}{19.5}     & \multicolumn{1}{c|}{30.7}      & \multicolumn{1}{c|}{27.8}        & \multicolumn{1}{c|}{24.5}  & \multicolumn{1}{c|}{60.7}  \\ \hline
\rowcolor{tableGray}
\multicolumn{1}{|l|}{RenderCNN \cite{Su_2015_ICCV}}        & \multicolumn{1}{c|}{27.5}         & \multicolumn{1}{c|}{25.8}     & \multicolumn{1}{c|}{6.5}     & \multicolumn{1}{c|}{45.8}    & \multicolumn{1}{c|}{29.7}    & \multicolumn{1}{c|}{8.5}      & \multicolumn{1}{c|}{12.0}      & \multicolumn{1}{c|}{31.4}          & \multicolumn{1}{c|}{17.7}     & \multicolumn{1}{c|}{29.7}      & \multicolumn{1}{c|}{31.4}        & \multicolumn{1}{c|}{24.2}  & \multicolumn{1}{c|}{56.9}  \\ \hline
\multicolumn{1}{|l|}{VP\&KP \cite{Tulsiani_vpsKps_2015}}        & \multicolumn{1}{c|}{46.6}         & \multicolumn{1}{c|}{42.0}     & \multicolumn{1}{c|}{12.7}     & \multicolumn{1}{c|}{64.6}    & \multicolumn{1}{c|}{42.7}    & \multicolumn{1}{c|}{20.8}      & \multicolumn{1}{c|}{18.5}      & \multicolumn{1}{c|}{38.8}          & \multicolumn{1}{c|}{33.5}     & \multicolumn{1}{c|}{42.5}      & \multicolumn{1}{c|}{32.9}        & \multicolumn{1}{c|}{\textbf{36.0}}  & \multicolumn{1}{c|}{56.9}   \\ \hline
\multicolumn{1}{|l|}{Specific-network}        & \multicolumn{1}{c|}{36.7}         & \multicolumn{1}{c|}{30.5}     & \multicolumn{1}{c|}{11.7}     & \multicolumn{1}{c|}{57.4}    & \multicolumn{1}{c|}{39.7}    & \multicolumn{1}{c|}{8.9}      & \multicolumn{1}{c|}{21.8}      & \multicolumn{1}{c|}{29.6}          & \multicolumn{1}{c|}{25.5}     & \multicolumn{1}{c|}{38.0}      & \multicolumn{1}{c|}{31.9}        & \multicolumn{1}{c|}{30.2} & \multicolumn{1}{c|}{\textbf{64.3}}   \\ \hline 
\multicolumn{14}{c}{AVP 24 Views - Joint Object Detection and Pose Estimation}                                                                                                                                                                                                                                                                                                              \\ \hline
\rowcolor{tableGray}
\multicolumn{1}{|l|}{VDPM \cite{xiang2014}}         & \multicolumn{1}{c|}{8.0}     & \multicolumn{1}{c|}{14.3}     & \multicolumn{1}{c|}{0.3}    & \multicolumn{1}{c|}{39.2}    & \multicolumn{1}{c|}{13.7}      & \multicolumn{1}{c|}{4.4}      & \multicolumn{1}{c|}{3.6}          & \multicolumn{1}{c|}{10.1}     & \multicolumn{1}{c|}{8.2}      & \multicolumn{1}{c|}{20.0}       
& \multicolumn{1}{c|}{11.2}        & \multicolumn{1}{c|}{12.1}  & \multicolumn{1}{c|}{29.5}  \\ \hline
\rowcolor{tableGray}
\multicolumn{1}{|l|}{DPM-VOC+VP \cite{Pepik2012}}        & \multicolumn{1}{c|}{9.7}         & \multicolumn{1}{c|}{16.7}     & \multicolumn{1}{c|}{2.2}     & \multicolumn{1}{c|}{42.1}    & \multicolumn{1}{c|}{24.6}    & \multicolumn{1}{c|}{4.2}      & \multicolumn{1}{c|}{2.1}      & \multicolumn{1}{c|}{10.5}          & \multicolumn{1}{c|}{4.1}     & \multicolumn{1}{c|}{20.7}      & \multicolumn{1}{c|}{12.9}        & \multicolumn{1}{c|}{13.6}   & \multicolumn{1}{c|}{27.1}  \\ \hline
\rowcolor{tableGray}
\multicolumn{1}{|l|}{Craft-CNN \cite{MassaBMVC2016}}        &  \multicolumn{1}{c|}{42.4}     & \multicolumn{1}{c|}{37.0}    & \multicolumn{1}{c|}{18.0}    & \multicolumn{1}{c|}{59.6}      & \multicolumn{1}{c|}{43.3}      & \multicolumn{1}{c|}{7.6}          & \multicolumn{1}{c|}{25.1}     & \multicolumn{1}{c|}{39.3}      & \multicolumn{1}{c|}{29.4}        & \multicolumn{1}{c|}{48.1}     &  
\multicolumn{1}{c|}{28.4}         & \multicolumn{1}{c|}{\textbf{34.4}} & \multicolumn{1}{c|}{59.9}	\\ \hline
\rowcolor{tableGray}
\multicolumn{1}{|l|}{Single-Shot \cite{poirson2016_singleshot}}        & \multicolumn{1}{c|}{33.4}         & \multicolumn{1}{c|}{29.4}     & \multicolumn{1}{c|}{9.2}     & \multicolumn{1}{c|}{54.7}    & \multicolumn{1}{c|}{35.7}    & \multicolumn{1}{c|}{5.5}      & \multicolumn{1}{c|}{23.0}      & \multicolumn{1}{c|}{30.3}          & \multicolumn{1}{c|}{27.6}     & \multicolumn{1}{c|}{44.1}      & \multicolumn{1}{c|}{34.3}        & \multicolumn{1}{c|}{28.8} & \multicolumn{1}{c|}{59.3}   \\ \hline
\rowcolor{tableGray}
\multicolumn{1}{|l|}{SubCNN \cite{xiang2016subcategory}}        & \multicolumn{1}{c|}{20.7}         & \multicolumn{1}{c|}{16.4}     & \multicolumn{1}{c|}{7.9}     & \multicolumn{1}{c|}{34.6}    & \multicolumn{1}{c|}{24.6}    & \multicolumn{1}{c|}{9.4}      & \multicolumn{1}{c|}{7.6}      & \multicolumn{1}{c|}{19.9}          & \multicolumn{1}{c|}{20.0}     & \multicolumn{1}{c|}{32.7}      & \multicolumn{1}{c|}{18.2}        & \multicolumn{1}{c|}{19.3} & \multicolumn{1}{c|}{60.7}   \\ \hline
\rowcolor{tableGray}
\multicolumn{1}{|l|}{RenderCNN \cite{Su_2015_ICCV}}        & \multicolumn{1}{c|}{21.5}         & \multicolumn{1}{c|}{22.0}     & \multicolumn{1}{c|}{4.1}     & \multicolumn{1}{c|}{38.6}    & \multicolumn{1}{c|}{25.5}    & \multicolumn{1}{c|}{7.4}      & \multicolumn{1}{c|}{11.0}      & \multicolumn{1}{c|}{24.4}          & \multicolumn{1}{c|}{15.0}     & \multicolumn{1}{c|}{28.0}      & \multicolumn{1}{c|}{19.8}        & \multicolumn{1}{c|}{19.8}  & \multicolumn{1}{c|}{56.9}  \\ \hline
\multicolumn{1}{|l|}{VP\&KP \cite{Tulsiani_vpsKps_2015}}        & \multicolumn{1}{c|}{37.0}         & \multicolumn{1}{c|}{33.4}     & \multicolumn{1}{c|}{10.0}     & \multicolumn{1}{c|}{54.1}    & \multicolumn{1}{c|}{40.0}    & \multicolumn{1}{c|}{17.5}      & \multicolumn{1}{c|}{19.9}      & \multicolumn{1}{c|}{34.3}          & \multicolumn{1}{c|}{28.9}     & \multicolumn{1}{c|}{43.9}      & \multicolumn{1}{c|}{22.7}        & \multicolumn{1}{c|}{31.1}  & \multicolumn{1}{c|}{56.9}   \\ \hline
\multicolumn{1}{|l|}{Specific-network}        & \multicolumn{1}{c|}{22.9}         & \multicolumn{1}{c|}{21.8}     & \multicolumn{1}{c|}{8.8}     & \multicolumn{1}{c|}{45.0}    & \multicolumn{1}{c|}{33.2}    & \multicolumn{1}{c|}{7.0}      & \multicolumn{1}{c|}{18.2}      & \multicolumn{1}{c|}{20.8}          & \multicolumn{1}{c|}{16.9}     & \multicolumn{1}{c|}{33.4}      & \multicolumn{1}{c|}{21.8}        & \multicolumn{1}{c|}{22.7}  & \multicolumn{1}{c|}{\textbf{64.3}}  \\ \hline 
\end{tabular}
} 
\caption{Comparison with the state-of-the-art  in the PASCAL3D+ dataset.In gray color we show the \emph{joint} solutions.}
\label{tab:pascal3d_state-of-the-art}
\end{table*}

We now compare our best model, \ie the \textit{Specific-network}, with the state-of-the-art models in Table \ref{tab:pascal3d_state-of-the-art}. First of all, our \textit{Specific-network} reports the best object detection results: see last column in Table \ref{tab:pascal3d_state-of-the-art}. 

Depending on the the number set of views used for the evaluation in the PASCAL3D+ we can identify different winners, even from different families of methods. For instance, \emph{joint}  models retrieve the best results, in terms of mAVP, for 4, 8 and 24 views. For 16 views, it is the \emph{independent} model in \cite{Tulsiani_vpsKps_2015} the one reporting the best performance.

Regarding all the results in Table \ref{tab:pascal3d_state-of-the-art} we can conclude that the \emph{independent} approaches exhibit a better accuracy over most of the \emph{joint} models.

Note that the state-of-the-art for 24 view sets if achieved by the Craft-CNN \cite{MassaBMVC2016}, which uses synthetic CAD models during learning. This is also the case for the RenderCNN \cite{Su_2015_ICCV}. The rest of models, including ours, do not use any extra data in form of CAD models. Note that the \textit{Specific-network} systematically reports a better performance than the RenderCNN, for instance. The Single-Shot approach \cite{poirson2016_singleshot} is the clear winner for 4 and 8 set of views, and the VP\&KP \cite{Tulsiani_vpsKps_2015} wins for 16 set of views. In all these scenarios, our \textit{Specific-network} reports a higher detection accuracy than the winner model. This aspect is relevant, because the metric used tends to favor detectors with a lower localization precision. We refer the reader to the study in \cite{redondoCabrera2016} for more details. In other words, the more detections that are retrieved by a model, the greater the likelihood that the objects for which pose estimations have to be assigned are objects that, being more difficult to detect, appear occluded or truncated, or that are too small, aspects which naturally complicate a correct estimation of the viewpoint.

Every model comes with its own detector: VP\&KP uses the R-CNN \cite{girshick2014rcnn}, Craft-CNN uses the Fast R-CNN \cite{girshickICCV15fastrcnn}, and we follow the Faster R-CNN architecture \cite{ren2015fasterrcnn}. How can we evaluate the actual influence of the detector in the viewpoint estimation performance? In order to shed some light on this issue, we have decided to perform an additional experiment. We have taken the code of the VP\&KP model provided by the authors. This model defines an \emph{independent} type architecture, where two completely decoupled and different deep networks are used: one for detection, and one for the pose estimation. We start using our \textit{Specific-path} model which has the best detection performance, and we run it over the training images. We then collect these detections on the training data to enrich the ground truth data. Note that we only collect those detections whose overlap with the original ground truth is grater than 70\%. This is equivalent to the jittering technique applied in the original paper but taking into account the bounding box distribution of the detector. With this \emph{extended} training data, we proceed to train the original pose estimator in \cite{Tulsiani_vpsKps_2015}. For the test images, we recover our detections, and apply the described pose estimator on them. We call this pipeline: Improved VP\&KP (Imp-VP\&KP). Technically, the detector of the original VP\&KP has been improved, using the Faster R-CNN now.

\begin{figure}
\centering
\includegraphics[width=1.0\linewidth]{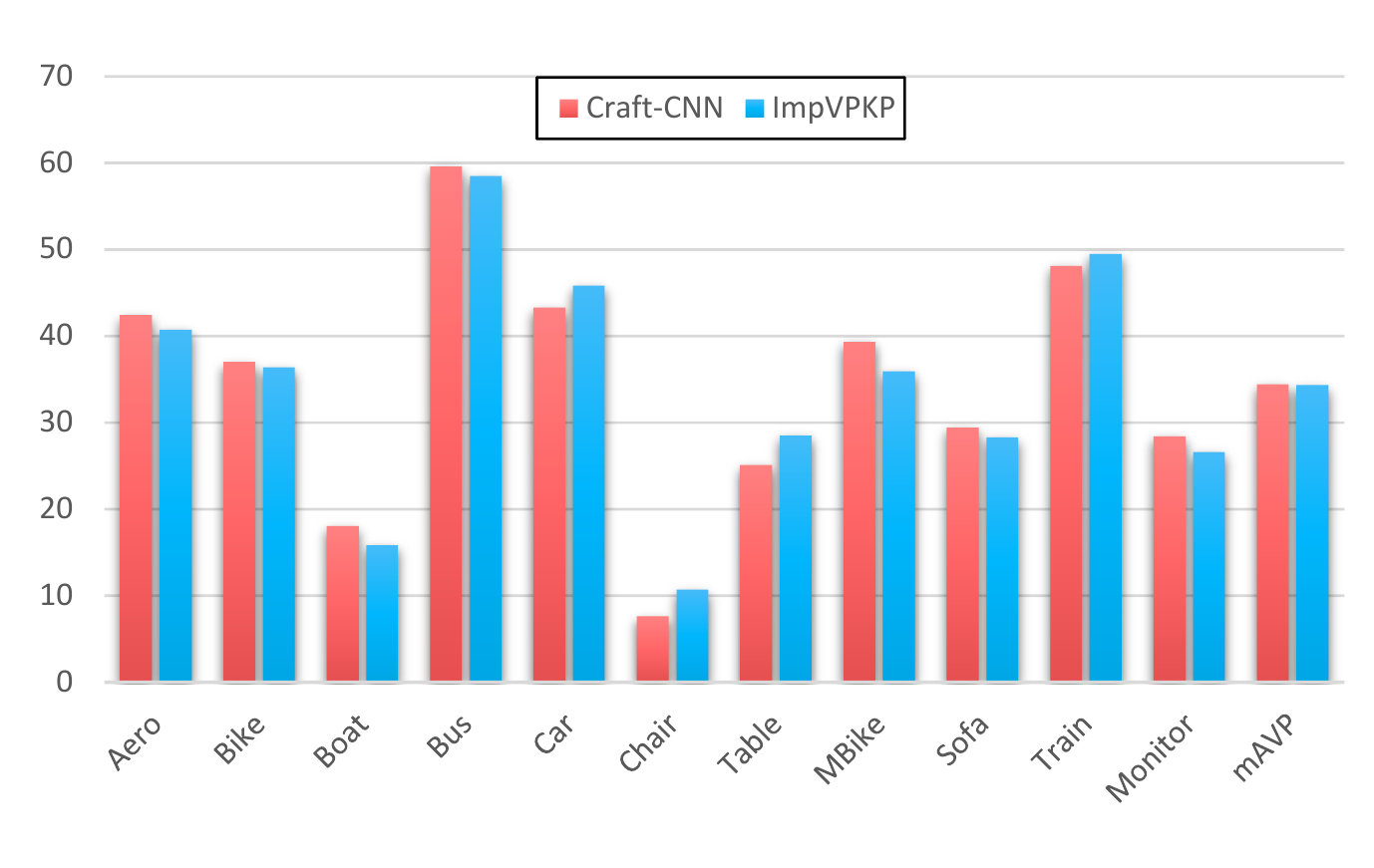} 
\caption{Comparison of Craft-CNN and the Imp-VP\&KP experiment for 24 views.}
\label{fig:craft-Imp-VPKP}
\end{figure}

As we can see in Table \ref{tab:Imp-VPKP_exp}, our Improved VP\&KP systematically reports better results than the original work. Moreover, in  Figure \ref{fig:craft-Imp-VPKP} we present a comparison between the Imp-VP\&KP and the results of the Craft-CNN \cite{MassaBMVC2016} for 24 views. We can observe how by simply updating the object detector, the model of \cite{Tulsiani_vpsKps_2015} can easily get the same performance as the Craft-CNN \cite{MassaBMVC2016}.

\begin{table*}[t]
\centering
\resizebox{0.8\linewidth}{!}{%
\begin{tabular}{lcccccccccccc}
\hline 
\multicolumn{1}{l}{Methods} & \multicolumn{1}{c}{Aero} & \multicolumn{1}{c}{Bike} & \multicolumn{1}{c}{Boat} & \multicolumn{1}{c}{Bus} & \multicolumn{1}{c}{Car} & \multicolumn{1}{c}{Chair} & \multicolumn{1}{c}{Table} & \multicolumn{1}{c}{MBike} & \multicolumn{1}{c}{Sofa} & \multicolumn{1}{c}{Train} & \multicolumn{1}{c}{Monitor} & \multicolumn{1}{c}{Avg.} \\ \hline 
\multicolumn{13}{c}{AVP 4 Views - Joint Object Detection and Pose Estimation}                                                                                                                                                                                                                                                                                                              \\ \hline
\multicolumn{1}{|l|}{VP\&KP \cite{Tulsiani_vpsKps_2015}}        & \multicolumn{1}{c|}{63.1}         & \multicolumn{1}{c|}{59.4}     & \multicolumn{1}{c|}{20.3}     & \multicolumn{1}{c|}{69.8}    & \multicolumn{1}{c|}{55.2}    & \multicolumn{1}{c|}{25.1}      & \multicolumn{1}{c|}{24.3}      & \multicolumn{1}{c|}{61.1}          & \multicolumn{1}{c|}{43.8}     & \multicolumn{1}{c|}{59.4}      & \multicolumn{1}{c|}{55.4}        & \multicolumn{1}{c|}{49.1}     \\ \hline

\multicolumn{1}{|l|}{Imp-VP\&KP}        & \multicolumn{1}{c|}{70.8}         & \multicolumn{1}{c|}{66.2}     & \multicolumn{1}{c|}{37.9}     & \multicolumn{1}{c|}{75.5}    & \multicolumn{1}{c|}{61.6}    & \multicolumn{1}{c|}{17.7}      & \multicolumn{1}{c|}{39.5}      & \multicolumn{1}{c|}{68.9}          & \multicolumn{1}{c|}{49.6}     & \multicolumn{1}{c|}{67.0}      & \multicolumn{1}{c|}{62.8}        & \multicolumn{1}{c|}{56.1}     \\ \hline
\multicolumn{13}{c}{AVP 8 Views - Joint Object Detection and Pose Estimation}                                                                                                                                                                                                                                                                                                             \\ \hline
\multicolumn{1}{|l|}{VP\&KP \cite{Tulsiani_vpsKps_2015}}        & \multicolumn{1}{c|}{57.5}         & \multicolumn{1}{c|}{54.8}     & \multicolumn{1}{c|}{18.9}     & \multicolumn{1}{c|}{59.4}    & \multicolumn{1}{c|}{51.5}    & \multicolumn{1}{c|}{24.7}      & \multicolumn{1}{c|}{20.5}      & \multicolumn{1}{c|}{59.5}          & \multicolumn{1}{c|}{43.7}     & \multicolumn{1}{c|}{53.3}      & \multicolumn{1}{c|}{45.6}        & \multicolumn{1}{c|}{44.5}     \\ \hline

\multicolumn{1}{|l|}{Imp-VP\&KP}        & \multicolumn{1}{c|}{63.9}         & \multicolumn{1}{c|}{61.4}     & \multicolumn{1}{c|}{29.0}     & \multicolumn{1}{c|}{63.3}    & \multicolumn{1}{c|}{56.2}    & \multicolumn{1}{c|}{15.8}      & \multicolumn{1}{c|}{32.8}      & \multicolumn{1}{c|}{65.3}          & \multicolumn{1}{c|}{42.0}     & \multicolumn{1}{c|}{60.6}      & \multicolumn{1}{c|}{53.6}        & \multicolumn{1}{c|}{49.4}     \\ \hline
\multicolumn{13}{c}{AVP 16 Views - Joint Object Detection and Pose Estimation}                                                                                                                                                                                                                                                                                                              \\ \hline
\multicolumn{1}{|l|}{VP\&KP \cite{Tulsiani_vpsKps_2015}}        & \multicolumn{1}{c|}{46.6}         & \multicolumn{1}{c|}{42.0}     & \multicolumn{1}{c|}{12.7}     & \multicolumn{1}{c|}{64.6}    & \multicolumn{1}{c|}{42.7}    & \multicolumn{1}{c|}{20.8}      & \multicolumn{1}{c|}{18.5}      & \multicolumn{1}{c|}{38.8}          & \multicolumn{1}{c|}{33.5}     & \multicolumn{1}{c|}{42.5}      & \multicolumn{1}{c|}{32.9}        & \multicolumn{1}{c|}{36.0}     \\ \hline

\multicolumn{1}{|l|}{Imp-VP\&KP}        & \multicolumn{1}{c|}{51.2}         & \multicolumn{1}{c|}{43.2}     & \multicolumn{1}{c|}{20.4}     & \multicolumn{1}{c|}{68.9}    & \multicolumn{1}{c|}{47.3}    & \multicolumn{1}{c|}{17.7}      & \multicolumn{1}{c|}{30.1}      & \multicolumn{1}{c|}{40.8}          & \multicolumn{1}{c|}{36.5}     & \multicolumn{1}{c|}{44.7}      & \multicolumn{1}{c|}{38.9}        & \multicolumn{1}{c|}{39.6}     \\ \hline
\multicolumn{13}{c}{AVP 24 Views - Joint Object Detection and Pose Estimation}                                                                                                                                                                                                                                                                                                              \\ \hline
\multicolumn{1}{|l|}{VP\&KP \cite{Tulsiani_vpsKps_2015}}        & \multicolumn{1}{c|}{37.0}         & \multicolumn{1}{c|}{33.4}     & \multicolumn{1}{c|}{10.0}     & \multicolumn{1}{c|}{54.1}    & \multicolumn{1}{c|}{40.0}    & \multicolumn{1}{c|}{17.5}      & \multicolumn{1}{c|}{19.9}      & \multicolumn{1}{c|}{34.3}          & \multicolumn{1}{c|}{28.9}     & \multicolumn{1}{c|}{43.9}      & \multicolumn{1}{c|}{22.7}        & \multicolumn{1}{c|}{31.1}     \\ \hline

\multicolumn{1}{|l|}{Imp-VP\&KP}        & \multicolumn{1}{c|}{40.7}         & \multicolumn{1}{c|}{36.4}     & \multicolumn{1}{c|}{15.8}     & \multicolumn{1}{c|}{58.5}    & \multicolumn{1}{c|}{45.8}    & \multicolumn{1}{c|}{10.7}      & \multicolumn{1}{c|}{28.5}      & \multicolumn{1}{c|}{35.9}          & \multicolumn{1}{c|}{28.3}     & \multicolumn{1}{c|}{49.5}      & \multicolumn{1}{c|}{26.6}        & \multicolumn{1}{c|}{34.3}     \\ \hline
\end{tabular}
} 
\caption{VP\&KP \cite{Tulsiani_vpsKps_2015} vs. Imp-VP\&KP experiment.}
\label{tab:Imp-VPKP_exp}
\end{table*}

\subsubsection{The side effect of the pose estimation in the joint system}
\label{sec:sec:pose_side_effect}

The systems that address the object detection and pose estimation problems simultaneously, in principle, have multiple benefits, compared with the models that decouple both tasks. They are clearly more efficient, in terms of computational cost. Note that during training, for instance, both task are learned simultaneously.  Moreover, for a test image, the localization of the object, and the estimation of its pose is obtained at the same time, not needing to process the images with a complex pipeline consisting of a detector followed by a viewpoint estimator. In a joint system, most of the operations are shared between tasks.

In spite of these advantages, our experiments reveal that there is a trade-off between doing the object localization accurately and casting a precise estimation for the viewpoint. Ideally, a good detector should be invariant to the different poses of an object, \eg it should correctly localize frontal and rear views of cars. This would push the detection models to learn representations that are not adequate to discriminate between the different poses, being this what a good pose estimator should learn. 

In Table \ref{tab:pose_side_effect} we report some results that can help us to understand the mentioned trade-off. The first two rows show the results reported by Massa \etal \cite{MassaBMVC2016}. They show a comparison between their joint and independent approaches. Their independent solution clearly obtains a better performance for the object detection than the joint model, but also one can observe how the pose estimation precision, in terms of mAVP, decreases. 

It is also interesting to observe, in the last rows of Table \ref{tab:pose_side_effect}, how this trade-off between object detection and pose estimation performances also affects the model of Poirson \etal \cite{poirson2016_singleshot}. We can see that when they try to train their Single-Shot joint model to be more discriminative in terms of poses,\ie increasing the number of sets of views from 4 to 24, the object detection accuracy tends to decrease.

If we now analyze the performance reported by our solutions, from the \textit{Single-path} to the \textit{Specific-network}, we note that the detection performance slightly increases for our \emph{independent} models, but we are able to also report a better performance in terms of pose estimation. We explain this fact with the type of deep architectures we have proposed. Both the \textit{Specific-path} and the \textit{Specific-network} can not be categorized as truly independent models: we do not completely decouple the tasks of object localization and pose estimation. Ours is an exercise or relaxing the amount of shared information between these tasks, which defines a training process able to enforce the networks to learn representations that are adequate for both tasks.
 
\begin{table}
\centering
\resizebox{\linewidth}{!}{%
\begin{tabular}{|l|c|c|}
\hline
Method & mAP & mAVP \\ \hline
\multicolumn{3}{c}{Craft-CNN (AlexNet) \cite{MassaBMVC2016}} \\
\hline
Joint - 24 views & 48.6                      & 21.1                      \\ \hline
Independent - 24 views & 51.6                      & 20.5                      \\ \hline
\multicolumn{3}{c}{Ours} \\
\hline
Joint - 24 views (Single-path) & 63.6 & 18.9 \\ \hline
Independent - 24 views (Specific-path) & 64.9 & 22.1 \\ \hline
Independent - 24 views (Specific-network) & 64.3 & 22.7 \\ \hline
\multicolumn{3}{c}{Single-Shot \cite{poirson2016_singleshot}} \\
\hline
Joint - 4 views & 61.0 & 50.7 \\ \hline
Joint - 8 views & 60.4 & 45.1 \\ \hline
Joint - 16 views & 60.0 & 33.3 \\ \hline
Joint - 24 views & 59.3 & 28.8 \\ \hline
\end{tabular}
} 
\caption{Analysis of the trade-off between object detection and pose estimation performance.}
\label{tab:pose_side_effect}
\end{table}

\subsection{Results in the ObjectNet3D dataset}

\begin{figure*}[t]
\centering
\includegraphics[width=0.75\linewidth]{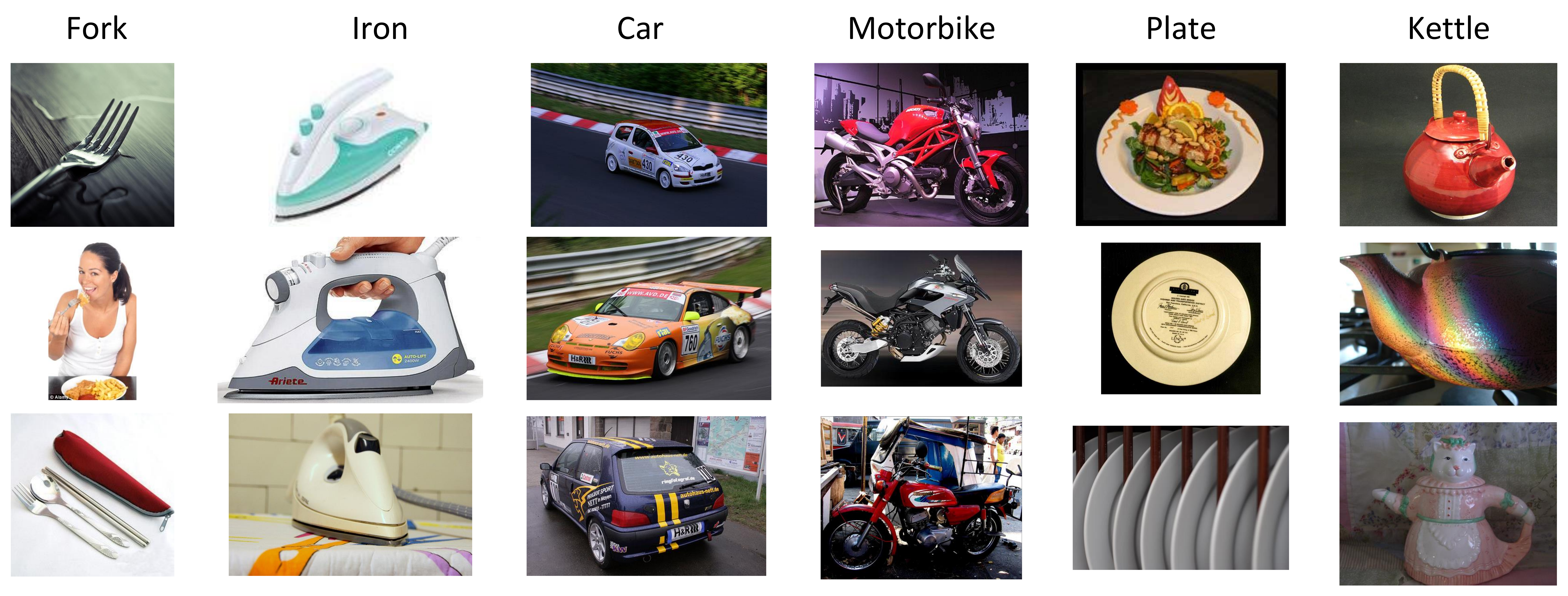} 
\caption{ObjectNet3D image samples.}
\label{fig:objectnet_sample}
\end{figure*}

In this work, we also perform a detailed experimental evaluation of our models in the large scale dataset for 3D object recognition ObjectNet3D \cite{xiang2016}. It consists of 100 categories, 90.127 images and more than 200.000 annotated objects. This dataset has been carefully designed for the evaluation of the problems of object detection, classification, and pose estimation. Similarly to the PASCAL3D+, the object pose annotation is the result of the manual alignment of a 3D CAD model with the target object. Figure \ref{fig:objectnet_sample} shows some examples of this dataset.

Like we describe in Section \ref{sec:implementation_details}, we strictly follow the experimental setup detailed in \cite{xiang2016}. Only the training data is used to learn the models. We then report our results using the validation and test sets. For the evaluation metric, Xiang \etal \cite{xiang2016} propose a generalization of the AVP. They basically extend the AVP to consider the prediction of the three angles provided in the annotation: azimuth, elevation and in-plane rotation. Technically, the solutions must provide an estimation for these three variables. Then, the corresponding predicted rotation matrix $\hat{R}$ is constructed. The difference between the ground truth pose $R$, and the prediction encoded in $\hat{R}$ is computed using a geodesic distance as follows:

\begin{equation}
d(R, \hat{R}) = \frac{1}{\sqrt{2}}|| \log \left( \hat{R}^T R \right)|| \, .
\end{equation}

According to Xiang \etal in \cite{xiang2016}, for the AVP, an estimation is considered to be correct if $d(R, \hat{R}) < \frac{\pi}{6}$. 

With respect to the technical implementation of our models, note that they cast a prediction for the three pose angles (azimuth, elevation and in-plane rotation) simultaneously. We repeat the same initialization procedure, using a pre-trained model on the ImageNet dataset. Again, we use the Stochastic Gradient Descent optimizer, with a momentum of 0.9, and the weight decay is set to 0.0005. This time we fix to 1 all the specific learning rates for each layer. The training is performed in an end-to-end fashion following the Faster R-CNN procedure \cite{ren2015fasterrcnn}.

\subsubsection{Discrete vs. Continuous approaches analysis}

In our experiments with the PASCAL3D+ dataset, one of the main conclusions obtained has been that the discrete pose estimation models, based on classifiers, give better results than continuous pose estimation models. When the number of training samples is not large enough, and the pose annotations are not well balanced, a discrete estimation model is generally the best option. Now, with the novel ObjectNet3D dataset, which provides more viewpoint annotations for more object categories, we have the opportunity to explore whether we can obtain a better performance for the continuous approaches.

We follow the same procedure described in Section \ref{sec:pascal_loss_analysis} for our previous Discrete vs. Continuous approaches analysis. We use our \textit{Single-path} model, which is trained for a continuous pose estimation task, solving a regression problem using the Euclidean and Huber losses. When the discrete pose estimation problem is tackled, we simply learn a classifier employing the Softmax loss.

\begin{table}
\centering
\begin{tabular}{|l|c|c|}
\hline
Losses & mAP & mAVP  \\
\hline\hline
Discrete (Eq. \ref{eq:softmax}) & 59.7 & 40.9 \\ \hline
Euclidean (Eq. \ref{eq:euclideanloss}) & 60.5 & 41.2 \\ \hline
Huber (Eq. \ref{eq:huberloss}) & 60.4 & 41.5  \\ \hline
\end{tabular}
\caption{Loss function analysis for the ObjectNet3D dataset. Object detection and viewpoint estimation performances are reported.}
\label{tab:objectnet_loss_function}
\end{table}

Table \ref{tab:objectnet_loss_function} reports the obtained results of our \textit{Single-path} architecture, trained on the training set, and evaluated over the validation set. In our experiments, we observe a similar performance among all the models, but this time the continuous pose estimation architectures exhibit a small advantage, like we have previously suggested. Therefore, for the rest of the experiments in this dataset, we use the continuous viewpoint architecture, employing the Huber loss.

\subsubsection{Comparison of our architectures}

\begin{figure}[h]
\centering
\includegraphics[width=0.9\columnwidth]{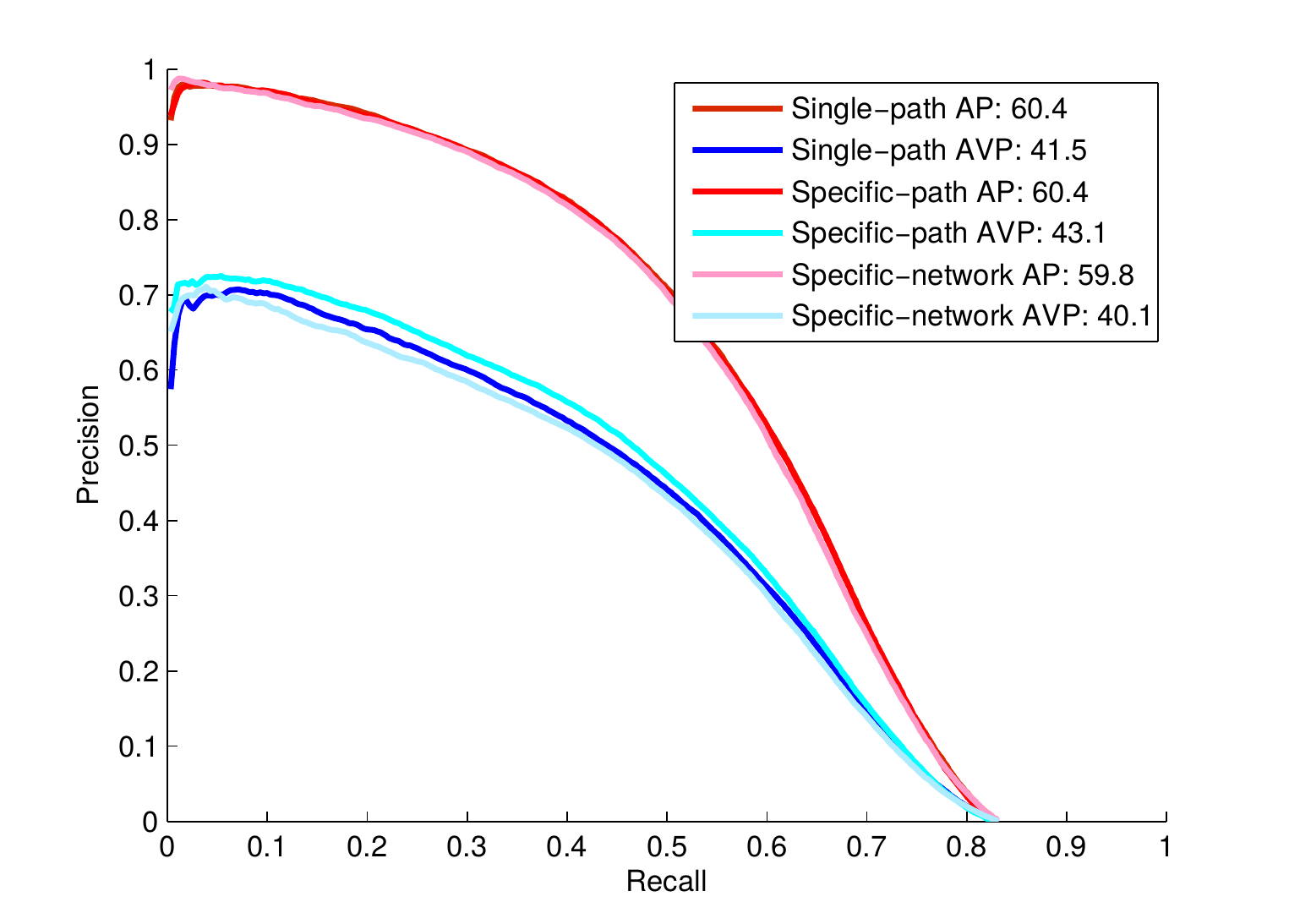} 
\caption{Object detection and pose estimation performance of our Single-path, Specific-path and Specific-network architectures in the ObjectNet3D dataset. Both AP and AVP metrics, with their associated precision-recall curves, are reported.}
\label{fig:objectnet_precision_recall}
\end{figure}

In this section, we propose to analyze the performance of all our architectures, \ie the Single-path, the Specific-path and the Specific-network, in this novel dataset. We only use the training set for learning the models, and the evaluation is carried in the validation set. Figure \ref{fig:objectnet_precision_recall} shows that for this dataset, all our models report a very similar performance. Note how the AP reported for the object localization task is almost identical for the three networks, while for the pose estimation the Specific-path exhibits a slightly superior AVP. In any case, we conclude that for this dataset, there is no clear winner within our models. Therefore, now that the amount of training data in the ObjectNet3D dataset has increased considerably, it seems that there are no major differences between treating the problem of locating objects and estimating their pose jointly or separately.

\subsubsection{A comparison with the State-of-the-art}

In this section, we provide a comparison with the state-of-the-art models reported by Xiang \etal \cite{xiang2016}. For the joint 2D detection and continuous 3D pose estimation task, they propose a modification of the Fast R-CNN \cite{girshickICCV15fastrcnn} model, using two different base architectures: the VGG-16 \cite{simonyan2014} and the AlexNet \cite{krizhevsky2012}. Technically, they add a viewpoint regression FC branch just after the FC7 layer. Their network is trained to jointly solve three tasks: classification, bounding box regression and viewpoint regression. The FC layer for viewpoint regression is of size $3 \times 101$, \ie, for each class, it predicts the three angles of azimuth, elevation and in-plane rotation. The smoothed L1 loss is used for viewpoint regression. 

\begin{table}
\centering
\begin{tabular}{|l|c|c|}
\hline
Method & mAP & mAVP \\
\hline
\hline
AlexNet \cite{xiang2016} & 54.2 & 35.4 \\ \hline
VGG-16 \cite{xiang2016} & \textbf{67.5} & 42.6 \\ \hline
Our & 64.2 & \textbf{46.7} \\ \hline
\end{tabular}
\caption {Comparison with state-of-the-art models in the ObjectNet3D dataset.}
\label{tab:objectnet_state-of-the-art}
\end{table}

Table \ref{tab:objectnet_state-of-the-art} shows the comparison with the state-of-the-art models, but now in the test set of the ObjectNet3D dataset. On the first two rows of the table, we include the results of the VGG-16 and the AlexNet based models reported in \cite{xiang2014}. The last row shows the performance of our \textit{Specific-path} model. First, note that we are able to report a better detection performance than the AlexNet based solution in \cite{xiang2014}. Second, although the VGG-16 based architecture of \cite{xiang2014} reports the best detection results, if we simultaneously consider the object localization and viewpoint estimation accuracies, \ie using the AVP metric, our model is the clear winner. This fact is particularly relevant, if we consider that the pose estimation is bounded by the detection performance, according to the evaluation metric used. Overall, this implies that our model is more accurate predicting poses.

In a detailed comparison of our solution with the VGG-16 based architecture used in \cite{xiang2014}, we find the following differences that also help to explain the results obtained. First, while our model is trained fully end-to-end, the approach in \cite{xiang2016} consists in training the Region Proposal Network of \cite{ren2015fasterrcnn} first, and then using these proposals to fine-tune their model for the object detection and pose estimation tasks. Therefore, their model is mainly trained to optimize the detection performance, which explains why our \textit{Specific-path} reports a slightly lower mAP. Second, there are significant differences in how the pose estimation is performed. In \cite{xiang2016}, a regressor is trained to directly predict viewpoint values in degrees. We, instead, decompose each angle into two polar coordinates. This decomposition naturally takes into account the cyclic nature of viewpoint angles. This explains why our \textit{Single-path} model reports a better performance for the pose.

\begin{figure*}[t]
\centering
\includegraphics[width=0.95\linewidth]{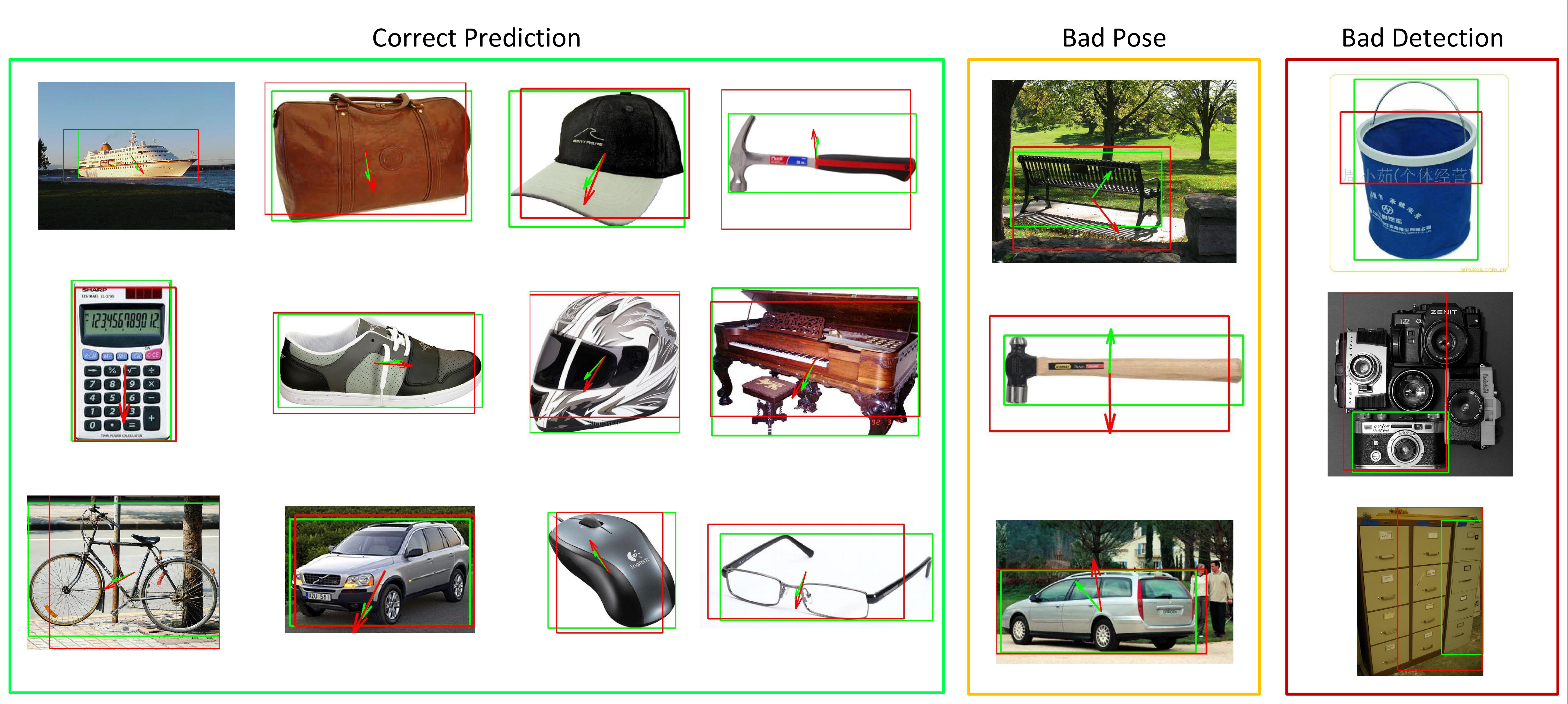} 
\caption{Qualitative results produced by the Specific-path in the ObjectNet3D. In green, we depict the ground truth annotations, while in red we show the results produced by our model. Rectangles correspond to the bounding boxes, while the arrows depict annotated orientations of the objects.}
\label{fig:objectnet_qualitative}
\end{figure*}

We finally show some qualitative results for the ObjectNet3D dataset in Figure \ref{fig:objectnet_qualitative}.

\section{Conclusion}
\label{sec:conclusions}

In this work, we have presented a complete analysis of the state of the art for the problem of simultaneous object detection and pose estimation. We have designed an experimental validation, using the PASCAL3D+ and ObjectNet3D datasets, where we can evaluate the degree of coupling that exists among the tasks of object localization and viewpoint estimation. For doing so, we have introduced three deep learning architectures, which are able to perform a joint detection and pose estimation, where we gradually decouple these two tasks. With the proposed models we have achieved the state-of-the-art performance in both datasets. We have concluded that decoupling the detection from the viewpoint estimation task have benefits on the overall performance of the models.

Furthermore, we have extended the comparative analysis of all our approaches considering the pose estimation as a discrete or a continuous problem, according to the two families of work in the literature. In our experiments, we have analyzed the main factors that need to be considered during the system design and training. Despite the similar performance among the different approaches, we have observed a difference between the discrete and the continuous models. We conclude that the continuous approaches are more sensitive to the pose bias in the annotation than the discrete models, hence requiring bigger datasets.

\section*{Acknowledgments}
This work is supported by project PREPEATE, with reference TEC2016-80326-R, of the Spanish Ministry of Economy, Industry and Competitiveness. We gratefully acknowledge the support of NVIDIA Corporation with the donation of the GPU used for this research. Cloud computing resources were kindly provided through a Microsoft Azure for Research Award.

\section*{References}

\bibliography{bib_collection}

\end{document}